\documentclass[10pt,twocolumn,letterpaper]{article}

\usepackage{wacv}              

\usepackage{graphicx}
\usepackage{amsmath}
\usepackage{amssymb}
\usepackage{booktabs}

\usepackage{color}
\usepackage{xspace}
\usepackage{xcolor}
\usepackage{float} 
\usepackage{multirow}
\usepackage{tcolorbox}
\usepackage{adjustbox}
\usepackage{pifont}
\usepackage{array}
\usepackage{makecell}
\usepackage{tabularray}
\UseTblrLibrary{booktabs}

\definecolor{wacvblue}{rgb}{0.21,0.49,0.74}
\usepackage[pagebackref,breaklinks,colorlinks,allcolors=wacvblue]{hyperref}

\usepackage[capitalize]{cleveref}
\crefname{section}{Sec.}{Secs.}
\Crefname{section}{Section}{Sections}
\Crefname{table}{Table}{Tables}
\crefname{table}{Tab.}{Tabs.}

\def\wacvPaperID{66} 
\def\confName{WACV}
\def\confYear{2026}

\begin{document}

\title{Improving Masked Style Transfer using Blended Partial Convolution}

\author{Seyed Hadi Seyed
\and
Ayberk Cansever\\
East Carolina University\\
1000 East Fifth Street, Greenville, NC, USA \\
*{\tt\small hartda23@ecu.edu}
\and
*David Hart\\
}
\maketitle

\begin{abstract}
   Artistic style transfer has long been possible with the advancements of convolution- and transformer-based neural networks. Most algorithms apply the artistic style transfer to the whole image, but individual users may only need to apply a style transfer to a specific region in the image. The standard practice is to simply mask the image after the stylization. This work shows that this approach tends to improperly capture the style features in the region of interest. We propose a partial-convolution-based style transfer network that accurately applies the style features exclusively to the region of interest. Additionally, we present network-internal blending techniques that account for imperfections in the region selection. We show that this visually and quantitatively improves stylization using examples from the SA-1B dataset. Code is publicly available at \href{https://github.com/davidmhart/StyleTransferMasked}{https://github.com/davidmhart/StyleTransferMasked}.
\end{abstract}

\section{Introduction}
\label{sec:intro}



Image style transfer aims to reproduce the aesthetic characteristics of a style image, such as a painting, within the structural context of a content image. This field has been deeply explored over the last decade \cite{gatys2016image,gatys2017controlling,li2017universal,li2019linear,deng2022stytr2}. These works, however, have focused on stylizing a whole content image based on a single style image.

Also in recent years, the Segment Anything Model \cite{kirillov2023sam} and follow-on work \cite{ma2024segmentmed,zou2024segmenteverything} have revolutionized segmentation tasks by allowing for a single click, non-class-based detector. Developments in this model are making segmentation of semantically meaningful regions more accessible than ever before. 

This work explores the use of style transfer when applied to segmented regions of a content image. Common image processing software such as Photoshop allow users to style transfer images with masked content, but most programs simply style transfer the whole image and alpha blend the masked content back onto the original image. This naive approach can fail to capture the proper style statistics due to the disparity between the distribution of colors in the whole image compared to the masked region.
Very few other works have focused on the problem of style transfer within a masked region \cite{Psychogyios2023SAMStylerEV} and none have presented a feed-forward model that does so.

\begin{figure}[t]
    \centering
    \begin{minipage}[b]{0.27\columnwidth}
        \centering
        \includegraphics[width=\textwidth,height=1.75cm]{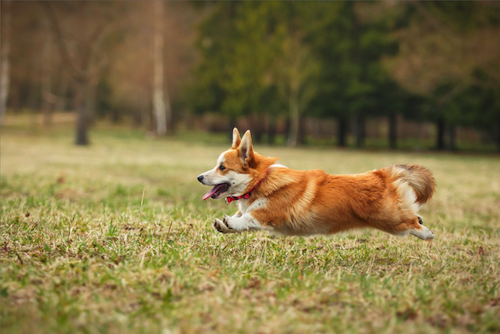}
    \end{minipage}
    \begin{minipage}[b]{0.16\columnwidth}
        \centering
        \includegraphics[width=0.6\textwidth,height=0.8cm]{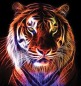}\\[1mm]
        \includegraphics[width=\textwidth,height=0.8cm]{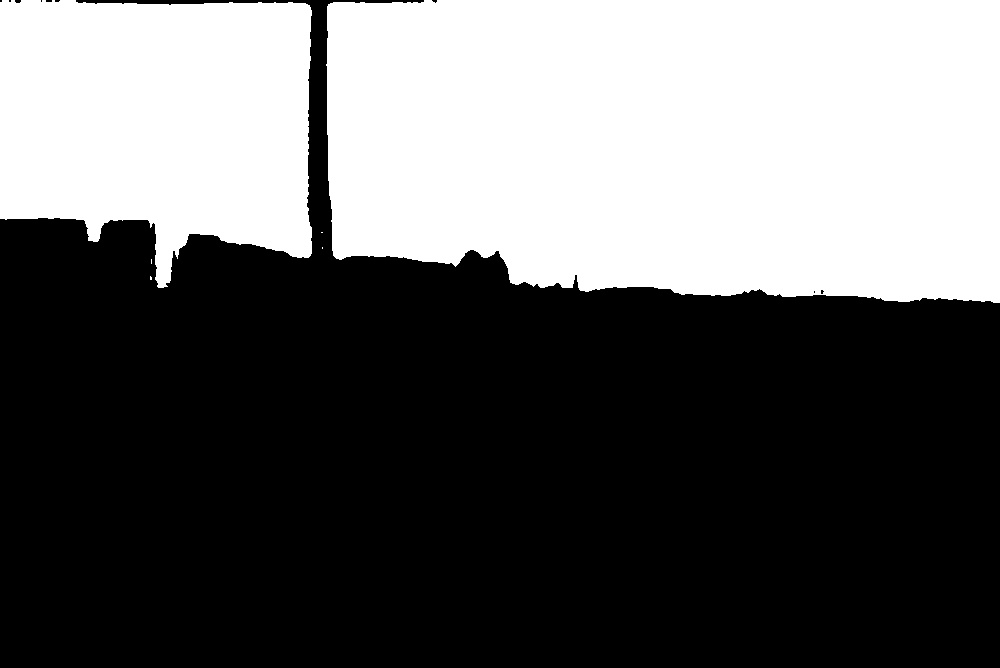}
    \end{minipage}
    \begin{minipage}[b]{0.27\columnwidth}
        \centering
        \includegraphics[width=\textwidth,height=1.75cm]{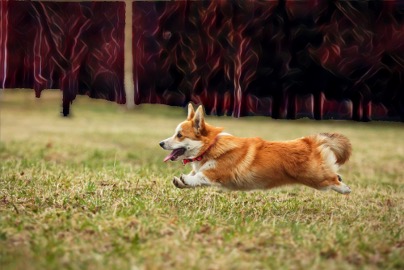}
    \end{minipage}
    \begin{minipage}[b]{0.27\columnwidth}
        \centering
        \includegraphics[width=\textwidth,height=1.75cm]{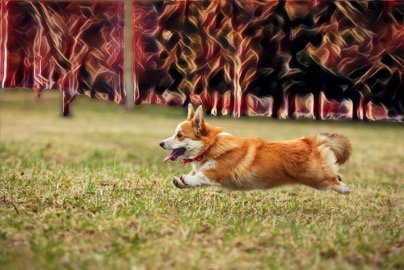}
    \end{minipage}
    \begin{minipage}[b]{0.27\columnwidth}
        \centering
        \includegraphics[width=\textwidth,height=1.75cm]{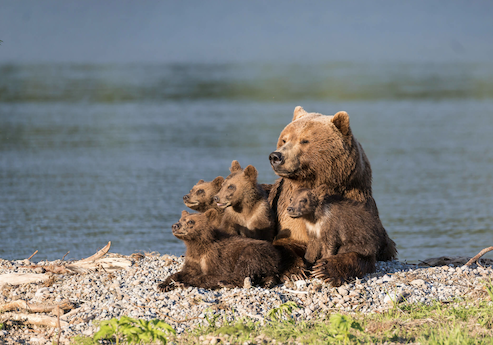}
    \end{minipage}
    \begin{minipage}[b]{0.16\columnwidth}
        \centering
        \includegraphics[width=0.6\textwidth,height=0.8cm]{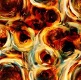}\\[1mm]
        \includegraphics[width=\textwidth,height=0.8cm]{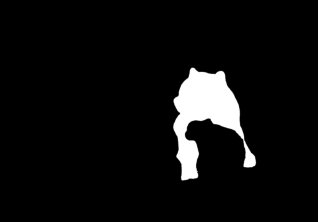}
    \end{minipage}
    \begin{minipage}[b]{0.27\columnwidth}
        \centering
        \includegraphics[width=\textwidth,height=1.75cm]{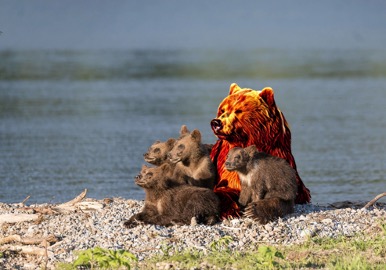}
    \end{minipage}
    \begin{minipage}[b]{0.27\columnwidth}
        \centering
        \includegraphics[width=\textwidth,height=1.75cm]{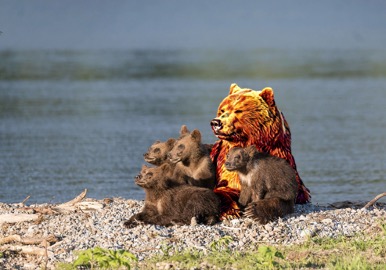}
    \end{minipage}
    \begin{minipage}[b]{0.27\columnwidth}
        \centering
        \includegraphics[width=\textwidth,height=1.75cm]{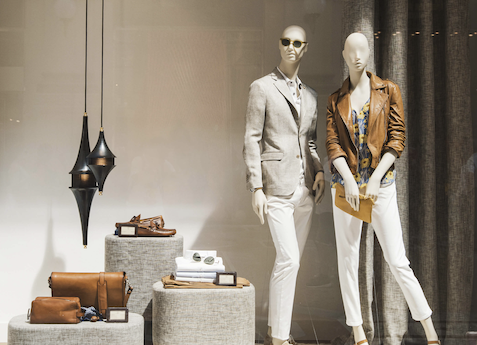}
    \end{minipage}
    \begin{minipage}[b]{0.16\columnwidth}
        \centering
        \includegraphics[width=0.6\textwidth,height=0.8cm]{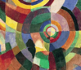}\\[1mm]
        \includegraphics[width=\textwidth,height=0.8cm]{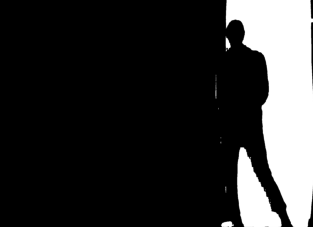}
    \end{minipage}
    \begin{minipage}[b]{0.27\columnwidth}
        \centering
        \includegraphics[width=\textwidth,height=1.75cm]{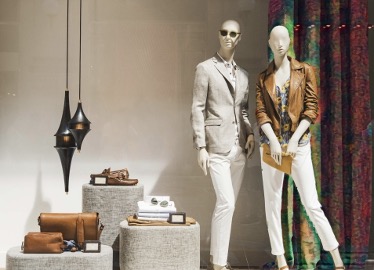}
    \end{minipage}
    \begin{minipage}[b]{0.27\columnwidth}
        \centering
        \includegraphics[width=\textwidth,height=1.75cm]{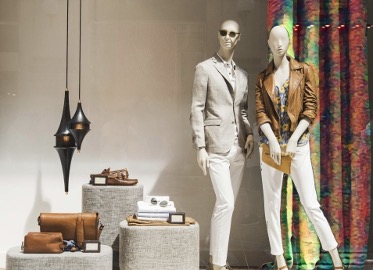}
    \end{minipage}   
    \vspace{2mm}
    \begin{minipage}[t]{0.27\columnwidth}
        \centering
        {\small a) Content} 
    \end{minipage}
    \begin{minipage}[t]{0.16\columnwidth}
        \centering
        {\small b) Style and Mask} 
    \end{minipage}
    \begin{minipage}[t]{0.27\columnwidth}
        \centering
        {\small c) Linear Style Transfer \cite{li2019linear}}
    \end{minipage}
    \begin{minipage}[t]{0.27\columnwidth}
        \centering
        {\small d) Ours}
    \end{minipage}

    \vspace{-3mm}
    \caption{
    Comparison of masked linear style transfer to our proposed partial-convolution-based approach. 
    When applying style transfer for a given content image (a), style, and mask (b), simply masking the output of the style transfer (c) does not provide well-stylized features in the masked region. Our approach that incorporates the mask throughout the style transfer process (d) better matches original style features in the masked region. 
    }
    \label{fig:comparison_style_then_mask_and_partial_conv}
\end{figure}

We present a feed-forward style transfer network that can style transfer any content and masked region given a particular style. This is done using partial convolutions \cite{liu2018partialconv} within the style transfer framework presented by Li \etal~\cite{li2019linear} without any additional fine-tuning. We will demonstrate that this approach better captures style properties within the masked region when compared to state-of-the-art methods that are masked after processing. Qualitatively, we verify this with many example outputs, such as those shown in Fig.~\ref{fig:comparison_style_then_mask_and_partial_conv}. Quantitatively, we show this by running metrics such as Earth Mover Distance and Perceptual Style Loss metrics against 500 images from SA-1B dataset.

In addition to these analyses, we provide many practical improvements to the stylization. First, although partial convolution proves effective for style transfer in masked regions, visible seams often appear where stylized and non-stylized areas meet. To address these border artifacts, we explore various blending techniques that smooth the border and help account for inaccuracies in the segmentation, producing more coherent and visually appealing results. Second, the partial convolution network naturally lends itself to stylizing multiple regions in parallel, combining features at a network level and creating smoothly transitioning results from one stylization to another. We describe how to setup such a mutli-mask, multi-style network and showcase its results.

In summary, our contributions include:

\begin{itemize}
    \item A new partial-convolution-based style transfer network that effectively stylizes masked content in an image without additional fine-tuning.
    \item Qualitative and quantitative results of the approach on 500 images from the SA-1B dataset.
    \item Blending techniques that improve the appearance of the stylization relative to the background.
    \item A multi-mask and multi-style configuration that stylizes in a single pass.
\end{itemize}

\section{Related Works}

\subsection{Style Transfer}
Modern style transfer techniques begin with the groundbreaking work of Gatys et al. \cite{gatys2016image} who presented the first CNN-based style transfer algorithm. This approach was optimization based and operated on a single image and single style at a time. Others works improved the speed and control of this approach \cite{gatys2017controlling, Kolkin2019, Puy2019}.
Later, a feed-forward method was presented by Johnson et al. \cite{johnson2016perceptual} that trained a network for a single style image. After this training process, any content image could be fed into the network and stylized in a quick feed-forward pass. Others also improved this approach \cite{Kotovenko2019, Ulyanov2017, ConditionalStyleTransfer}.
In 2017, style transfer was reformulated as a modified image reconstruction process \cite{li2017universal}. After a standard image reconstruction network is trained (usually an autoencoder), the content image is fed into the network and the intermediate representation is altered based on the style statistics. This approach is considered state-of-the-art and has CNN \cite{li2019linear}, vision transformer \cite{deng2022stytr2}, and even diffusion network \cite{zhang2023inversion,styleID} implementations.
In this work, the linear transformer apporach by Li \etal \cite{li2019linear} is used because it is a fully convolutional neural network, making it easy to modify the convolution layers to be partial convolutions.

\subsection{Segment Anything Models}
Segmentation is a common computer vision task. Before the Segment Anything Model (SAM)~\cite{kirillov2023sam}, modern AI-based approaches were class-based, relying on class labels to complete the segmentation \cite{long2015fully}. Popular approaches included models like Detectron \cite{wu2019detectron2}. The Segment Anything Model is unique because it was trained in an iterative fashion that does not require class labels, thus allowing it to segment objects with high accuracy, even if they have not been seen by the network before or have no semantic meaning. Many works have built on this approach \cite{ma2024segmentmed, zou2024segmenteverything}.

\subsection{Segmented Style Transfer}
Few works have yet to attempt to complete style transfer on segmented regions. Some approaches use segmentation to influence style transfer \cite{zhao2020automatic,lin2021image} or use class-based labels \cite{kurzman2019class, kulkarni2024improvedobjectbasedstyletransfer}. Recently, a work titled SAMStyler \cite{Psychogyios2023SAMStylerEV} has combined the Segment-Anything Model with Style Transfer techniques, but relies on the slow optimization-based approach of the original Gatys implementation. In this work, we will present an approach that completes classless style transfer on segmented regions in a fast, feed-forward way. 



\section{Flaws in Masking Standard Style Networks}\label{sec:justification}

Throughout this work, we will refer to three stylization techniques which use the pretrained stylization network by Li \etal \cite{li2019linear} as their base:
\begin{itemize}
    \item Style-then-mask: stylize the image, then mask to include only the region of interest.
    \item Mask-then-style: mask the region of interest, applying black pixels elsewhere, then stylize the image.
    \item PartialConv: our approach that uses partial convolutions \cite{liu2018partialconv} in each layer of the stylization network to apply calculations only to the region of interest.
\end{itemize}

\noindent We will describe the specifics of implementing our approach in Section~\ref{sec:Methodology}, but first, we provide intuition for why style-then-mask and mask-then-style have flaws that can lead to poor stylizations. These flaws generalize to any feed forward style network that applies masking as a post process step.

\subsection{Analyzing Color Distributions}

Stylization networks operate on the pixel colors for the content and style images, then perform some operation (usually in feature space) to give the content pixel colors the statistics of the style pixel colors \cite{li2017universal}. Thus, the color distributions in an image for both the content and style images play a vital role on the final output of the stylization.

\begin{figure}[t]
\begin{center}
       \begin{tabular}{cccc}
        \includegraphics[width=0.20\linewidth, height=0.6in]{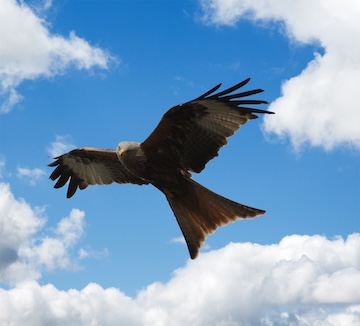}
        &
        \includegraphics[width=0.20\linewidth, height=0.6in]{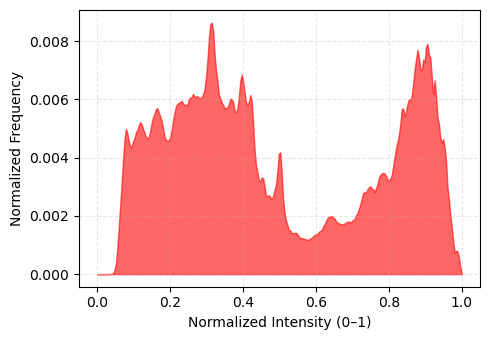}
        &
        \includegraphics[width=0.20\linewidth, height=0.6in]{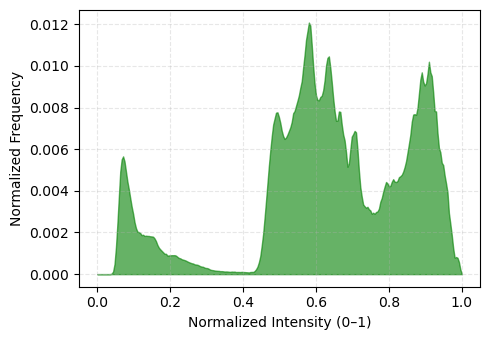}
        &
        \includegraphics[width=0.20\linewidth, height=0.6in]{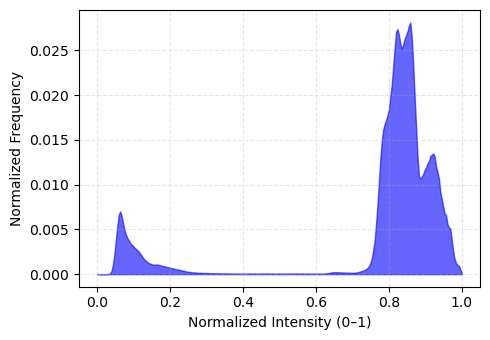}
        \\
        \includegraphics[width=0.20\linewidth, height=0.6in]{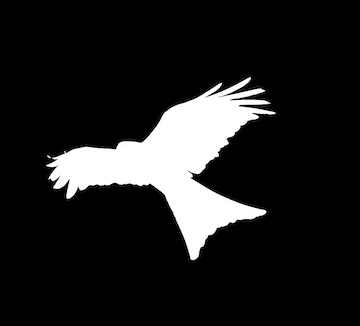}
        &
        \includegraphics[width=0.20\linewidth, height=0.6in]{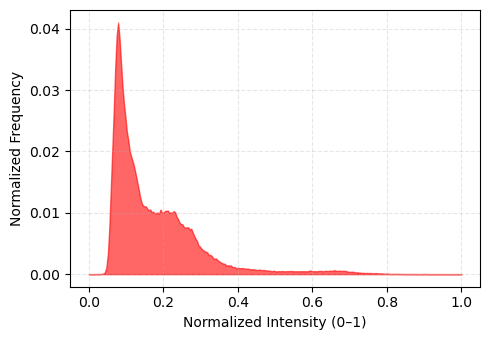}
        &
        \includegraphics[width=0.20\linewidth, height=0.6in]{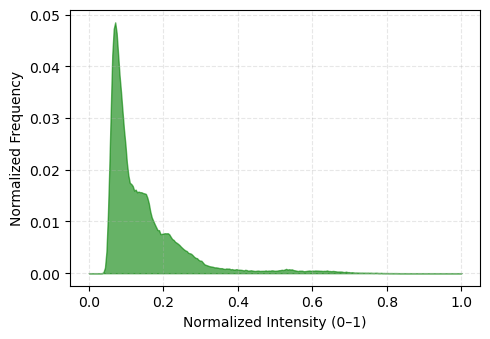}
        &
        \includegraphics[width=0.20\linewidth, height=0.6in]{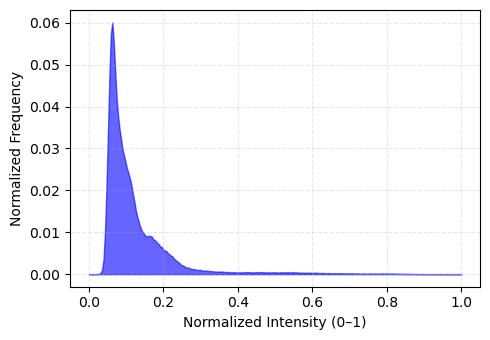}
    \end{tabular}
\end{center}
   \caption{The normalized histograms of pixel colors in red, green, and blue channels for the bird image. The statistics for original image (Top) and masked region (Bottom) are shown. The histograms show that the distribution of colors for the full image and masked region are disparate. This leads to the style-then-mask output misrepresenting aspects of the style in the masked region.}
\label{fig:distribution-bird}
\end{figure}

Simply styling an image and then masking to a region (style-then-mask) makes an important underlying assumption. It assumes that the color distributions for the whole image and the masked region are similar. This, however, is often not the case. The normalized histograms of the RGB colors for the content and its masked region of interest are not guaranteed to be the same, as shown in Fig.~\ref{fig:distribution-bird}.

\begin{figure}[t]
\begin{center}
        
       \begin{tabular}{cc}
        \includegraphics[width=0.45\linewidth]{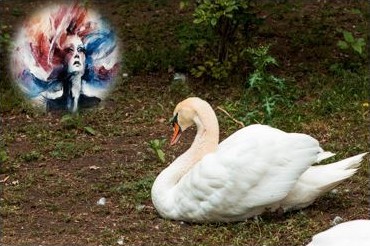}
        &
        \includegraphics[width=0.45\linewidth]{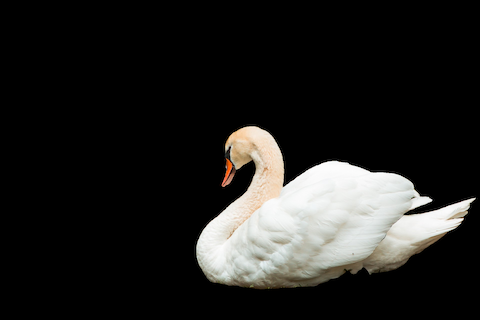}
        \\
        Original and Style &
        Masked Original
        \vspace{0.1in}
        \\
        \includegraphics[width=0.45\linewidth]{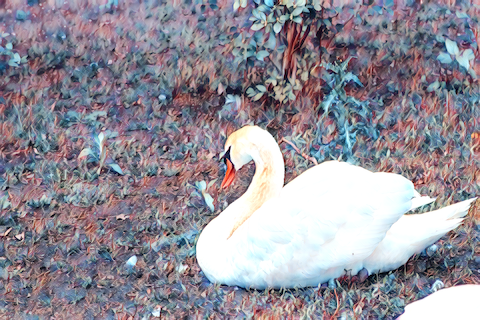}
        &
        \includegraphics[width=0.45\linewidth]{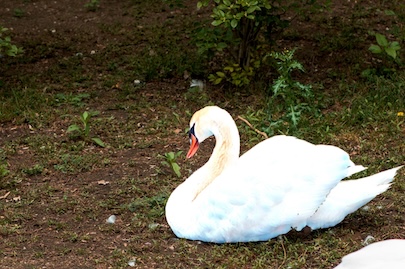}
        \\
        Stylized &
        Style-then-mask
        \vspace{0.1in}
        \\
        \includegraphics[width=0.45\linewidth]{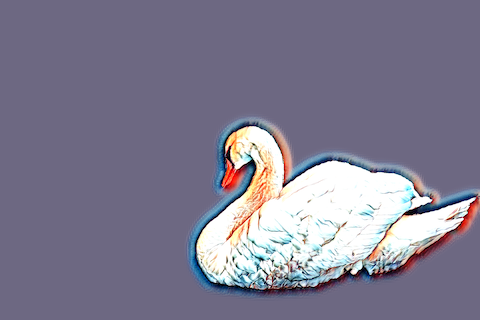}
        &
        \includegraphics[width=0.45\linewidth]{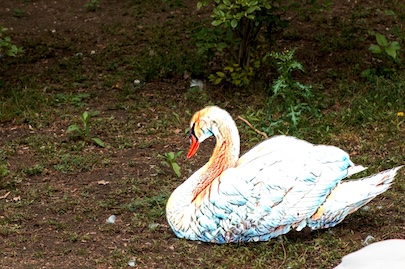}
        \\
        Stylized Masked Input &
        Mask-then-style
        \vspace{0.1in}
        \\
        \multicolumn{2}{c}{\includegraphics[width=0.45\linewidth]{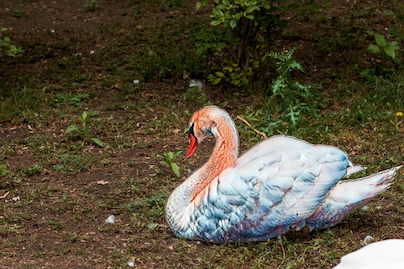}}
        \\
        \multicolumn{2}{c}{PartialConv}
        \\
    \end{tabular}
\end{center}
   \caption{An example comparing the three techniques: style-then-mask, mask-then-style, and partial convolution. A content image, style image, and mask (Row 1) give the shown stylized output for the masked swan regions. Stylizing the whole image and then masking out the region (Row 2) or masking out the region and then stylizing (Row 3) is compared to our approach (Row 4). The resulting stylization from partial convolution tends to be closer to the input style features than the other two approaches.}
\label{fig:partial-conv-better}
\end{figure}

Dissimilarities between the full image and masked region distributions cause issues in the stylization. Because the stylization networks apply style features across the whole image, this may under represent the distribution of style features and colors within the masked region. This can be visually seen in Fig. \ref{fig:partial-conv-better}. Style-then-mask and mask-then-style, which only perform a pre- or post-processing step, tend to have colors that are more washed out and overly dark or bright compared to the style. PartialConv, which only operates on pixels within the masked region, tends to much more closely match the true color distribution of the stylization image. As will be shown in Section ~\ref{sec:Results}, this can also be quantitatively verified with metrics such as Earth Movers Distance and Perceptual Style Loss.


\subsection{How often is this an issue?}


While images will almost always have disparate color distributions from the masked region of interest, the question naturally arises of how frequently the distributions are dissimilar enough to cause a noticeable change in the stylization. To answer this question, we conducted an experiment with 500 images from the SA-1B dataset \cite{kirillov2023sam} and 11 style images. For each content image and randomly selected style, a random mask was selected (with at least 2\% area of the image) and was stylized with the style-then-mask approach and our partial convolution approach. These two results and the original image, style, and mask were presented to a user. The user then indicated which method more closely matched the style features. They could also indicate that the two methods appeared the same. Mask-then-style was excluded from the experiment for its tendency to always over brighten the region of interest.

For these 500 images and masks, there were 212 times where PartialConv appeared to match the style features better than style-then-mask. In 283 cases, the two algorithms appeared to do the same and on rare occasion (5 times), style-then-mask appeared to do better. 

\newcommand{\styleimg}[1]{\raisebox{-.2\height}{\includegraphics[width=0.03\linewidth,height=0.03\linewidth]{#1}}}

\begin{table*}[t!]
    \centering
    \begin{tabular}{|c|ccc|ccc|}
        \hline
         & \multicolumn{3}{c|}{Same Output} & \multicolumn{3}{c|}{Improved with PartialConv} \\
         Style & Num Images & Gray EMD & Sliced EMD & Num Images & Gray EMD & Sliced EMD \\
         \hline
         \styleimg{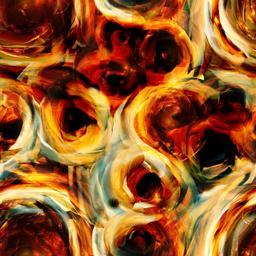} & 26 & 0.167 & 0.224 & 25 & \textbf{0.198} & \textbf{0.275} \\
         \styleimg{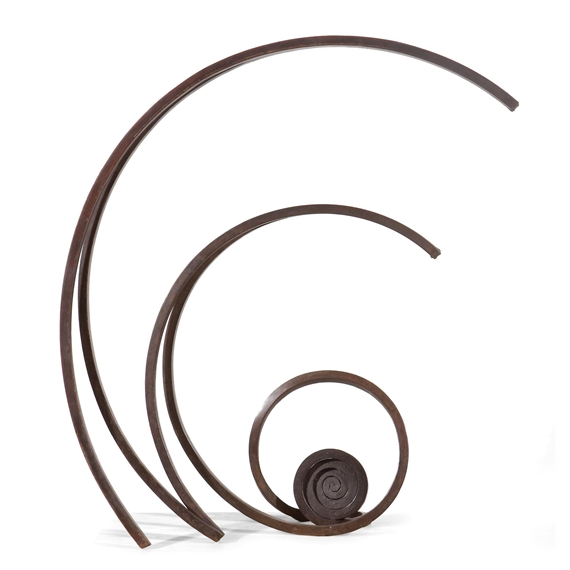} & 23 & \textbf{0.245} & \textbf{0.311} & 25 & 0.190 & 0.253  \\
         \styleimg{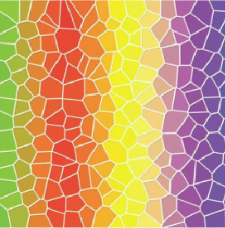} & 30 & 0.160 & 0.212 & 12 & \textbf{0.274} & \textbf{0.326}  \\
         \styleimg{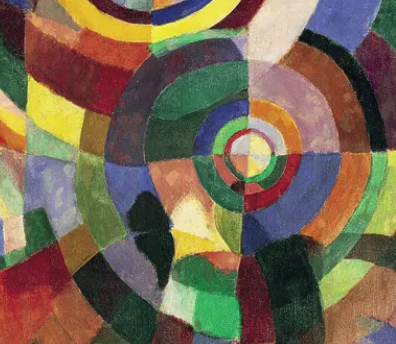} & 20 & 0.168 & 0.229 & 21 & \textbf{0.206} & \textbf{0.283}  \\
         \styleimg{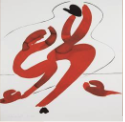} & 23 & 0.166 & 0.232 & 19 & \textbf{0.222} & \textbf{0.290}  \\
         \styleimg{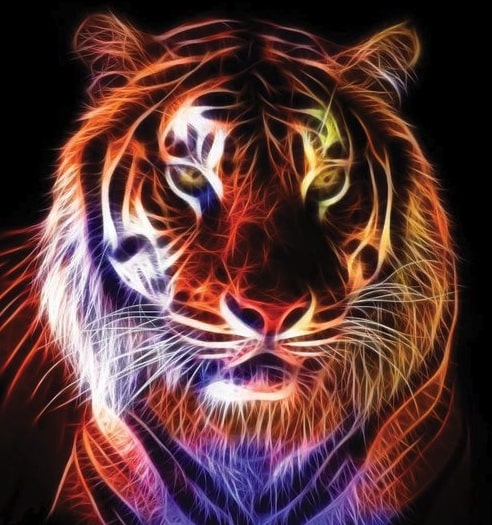} & 28 & 0.163 & 0.220 & 17 & \textbf{0.216} & \textbf{0.278}  \\
         \styleimg{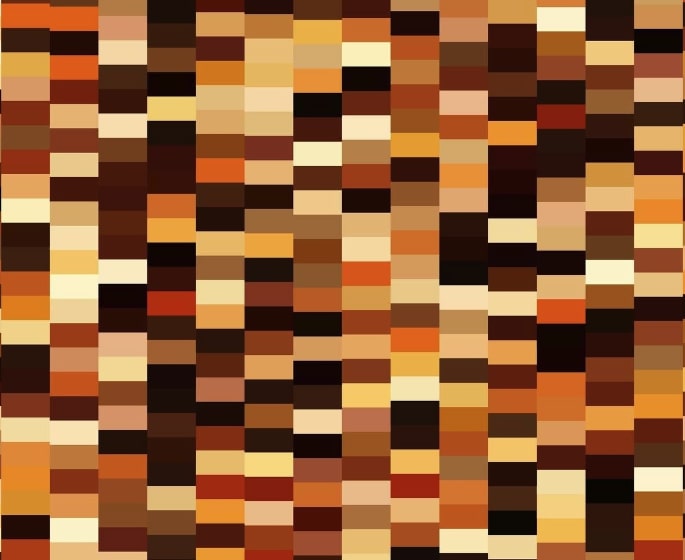} & 25 & 0.155 & 0.208 & 25 & \textbf{0.193} & \textbf{0.254}  \\
         \styleimg{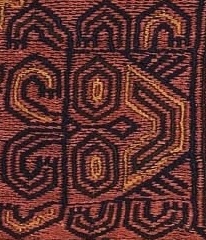} & 33 & \textbf{0.196} & 0.250 & 5 & 0.180 & \textbf{0.266}  \\
         \styleimg{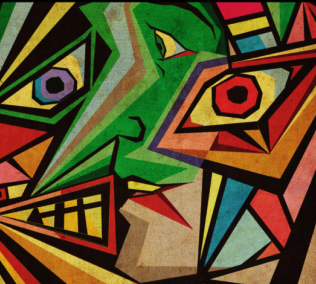} & 26 & 0.176 & 0.229 & 21 & \textbf{0.208} & \textbf{0.288}  \\
         \styleimg{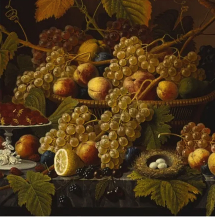} & 35 & \textbf{0.198} & \textbf{0.250} & 17 & 0.160 & 0.241  \\
         \styleimg{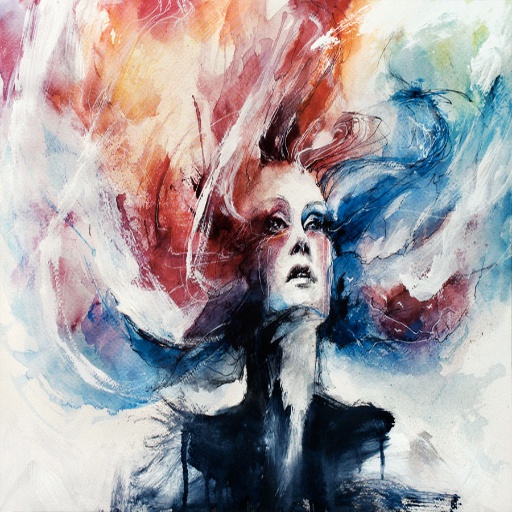} & 19 & 0.130 & 0.175 & 25 & \textbf{0.201} & \textbf{0.286} \\
         \hline
         Total/Average & 288 & 0.176 & 0.232 & 212 & \textbf{0.203} & \textbf{0.275}  \\
         \hline
    \end{tabular}
    \caption{The results of the visual study. For each content/mask/style that was presented to the user, the output of Style-then-mask and PartialConv were shown. The user then selected whether the outputs appeared the same, or if one output appeared to match the style statistics better. The statistics of the selection for each style is shown (since style-then-mask was selected as superior only 5 times, it is grouped with the `same' category). The Earth Movers Distance (EMD) between the whole image and masked area was also computed and averages are shown in the table. As can be seen, a larger EMD generally indicates that PartialConv will perform better, confirming the intuition that regular style transfer is insufficient the more dissimilar the whole-image and masked area color distributions become. The exceptions occur for styles that have narrow color distributions themselves.
    }
    \label{tab:EMD}
\end{table*}

For these 500 examples, the Earth Movers Distance (EMD) was processed between the whole image color distributions and the masked region. EMD assesses the alignment between the color distributions, with lower values indicating closer alignment. Rather than report red, green, and blue EMD, we summarize the EMD in two ways. First, we report EMD when computed on the grayscale version of the image. Second, we compute the sliced EMD that averages over projections of the 3D distribution in RGB. As expected, PartialConv tends to be a better stylization when the EMD is large between the masked and whole image, giving an average grayscale EMD of 0.203 and sliced EMD of 0.275. In comparison, the algorithms appeared similar when the EMD was lower, with an average grayscale EMD of 0.176 and sliced EMD of 0.232. The breakdown of EMD values for each style are shown in Tab. \ref{tab:EMD}.
These results showcase that EMD is an effective and intuitive metric for determining if masked style transfer will or will not benefit from partial convolution.



\section{Methodology}\label{sec:Methodology}

We now describe how to implement our approach of the partial convolution style network. We also describe some practical improvements that allow for better blending and multi-mask/multi-style options.

\subsection{Partial Convolution}\label{sec:PartialConv}

\begin{figure}
    \centering
    \includegraphics[width=0.98\linewidth]{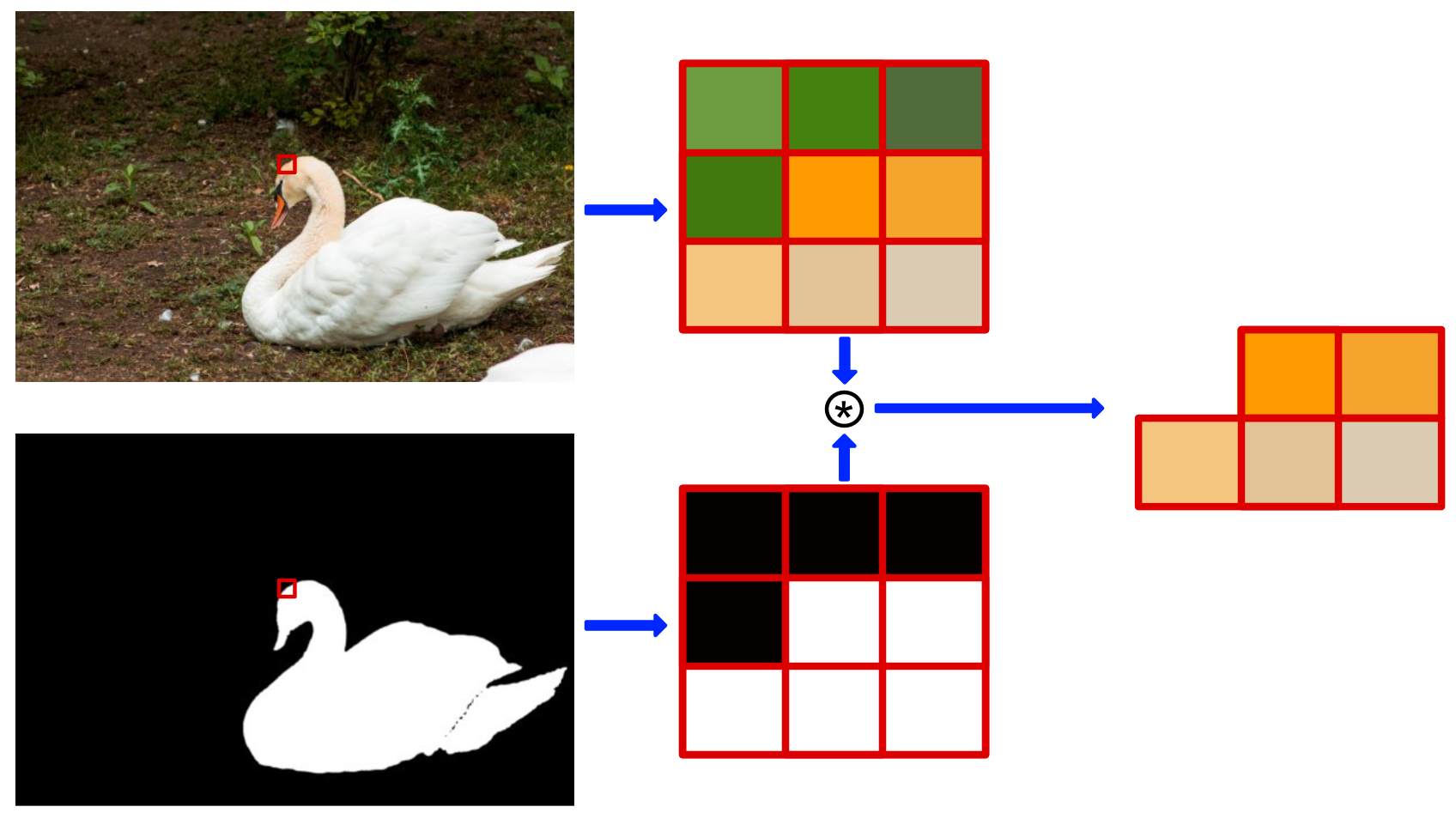}
    \caption{Visualization of the partial convolution operator. As the kernel for the CNN moves across the image, it is multiplied by the mask for the image, causing aggregation to only occur across pixels within the mask. This can effectively separate foreground information from background information.}
    \label{fig:partial_conv_operator}
\end{figure}

To set up the partial convolution style transfer network, the pretrained style transfer network presented by Li \etal is used \cite{li2019linear}. Every convolution layer in the encoder, decoder, and transformation blocks is replaced with partial convolutions \cite{liu2018partialconv}. Partial convolution allows for internal masking of the convolution operator in any neural network, only performing the weighted average over pixels within the mask, as illustrated in Fig.~\ref{fig:partial_conv_operator}. The same code and operator as presented in \cite{liu2018partialconv} is used, disabling the post convolution normalization that stabilizes training. Once each convolution in the style network is rewritten as a partial convolution, the same pretrained weights can be used without any additional fine-tuning. Our code for this modified style network is publicly available at \href{https://github.com/davidmhart/StyleTransferMasked}{https://github.com/davidmhart/StyleTransferMasked}.

Once the network is rewritten, the mask needs be properly accounted for at each stage of the style network. The input image and input mask are provided to the network for the computations. 
In the encoder, the mask goes through each padding and pooling layer that the
image goes through to guarantee the correct size of the mask at each layer. 
The matrix multiplication that performs the style transformation is also masked at
the feature level using the smallest output mask from the encoder. In the decoder, 
bilinear interpolation of the original mask is used to guarantee the correct size at each layer.
Finally, the output is alpha blended with the original image and mask to place the stylized region back onto the background. All of these steps modify the style transfer network to
only stylize the region inside the masked area without influencing the stylization with image
values outside the masked area.


\subsection{Blending}
This simple addition of partial convolution drastically improves stylization within the masked region, but masked style transfer also needs to be coherent with the background image it is placed on.
Effective blending is crucial to address artifacts in style application, especially around mask boundaries.
These artifacts arise from the variation in semantic region types and quality of the segmentation mask provided by SAM \cite{kirillov2023sam}. 

Several blending techniques were considered that operate on the mask during various stages of the network. The three blending techniques employed in our approach include:
\begin{itemize}
 \item \textbf{Mask Feathering Before Encoding:} Smoothes the edges of masks prior to encoding by gradually transitioning mask values from 1 to 0. 
 \item \textbf{Mask Expansion During Partial Convolution:} Dynamically dilates masks within convolutional layers, enabling convolution filters to access additional context around mask edges. 
 \item \textbf{Content Feathering in Decoder:} Includes the non-stylized content features from the standard autoencoder at each stage of the decoder, feathering the output mask of the style network to incorporate nearby content features. 
 
\end{itemize}

\begin{figure}[t]
    \centering

    \begin{tabular}{cc}
        \includegraphics[width=0.35\linewidth]{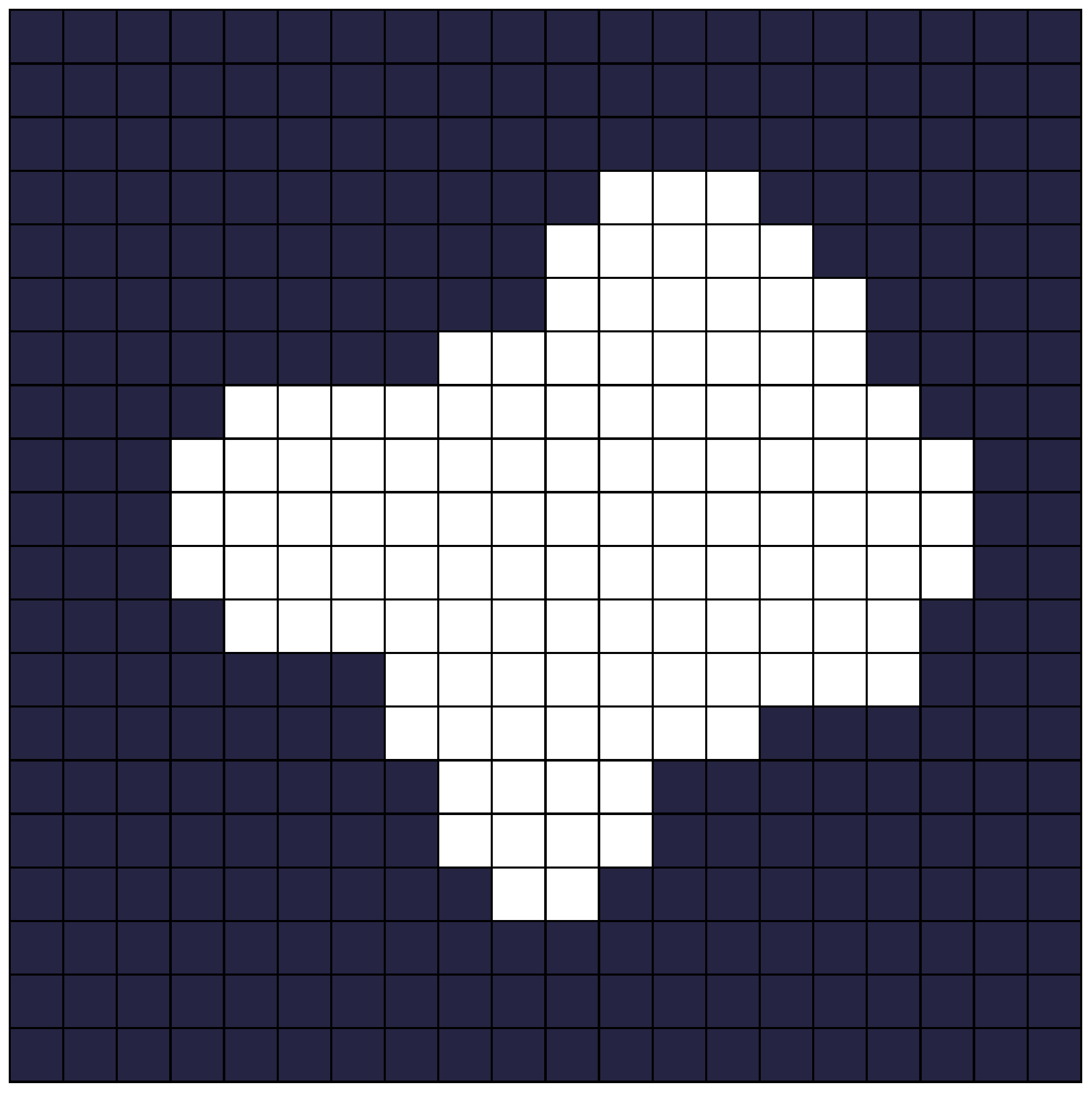} & \includegraphics[width=0.35\linewidth]{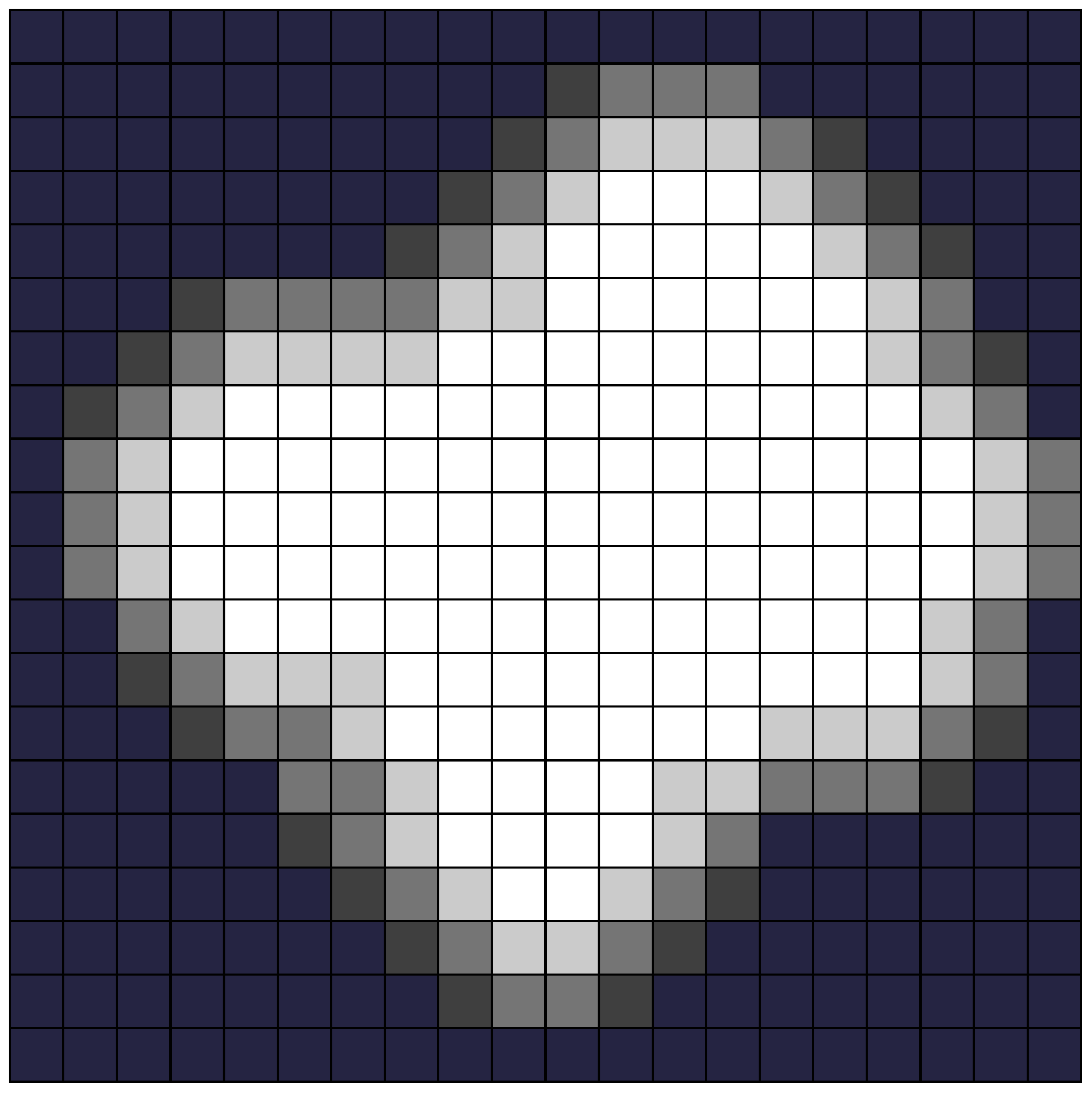}  \\
        Original & Mask Feathering \\
         \includegraphics[width=0.35\linewidth]{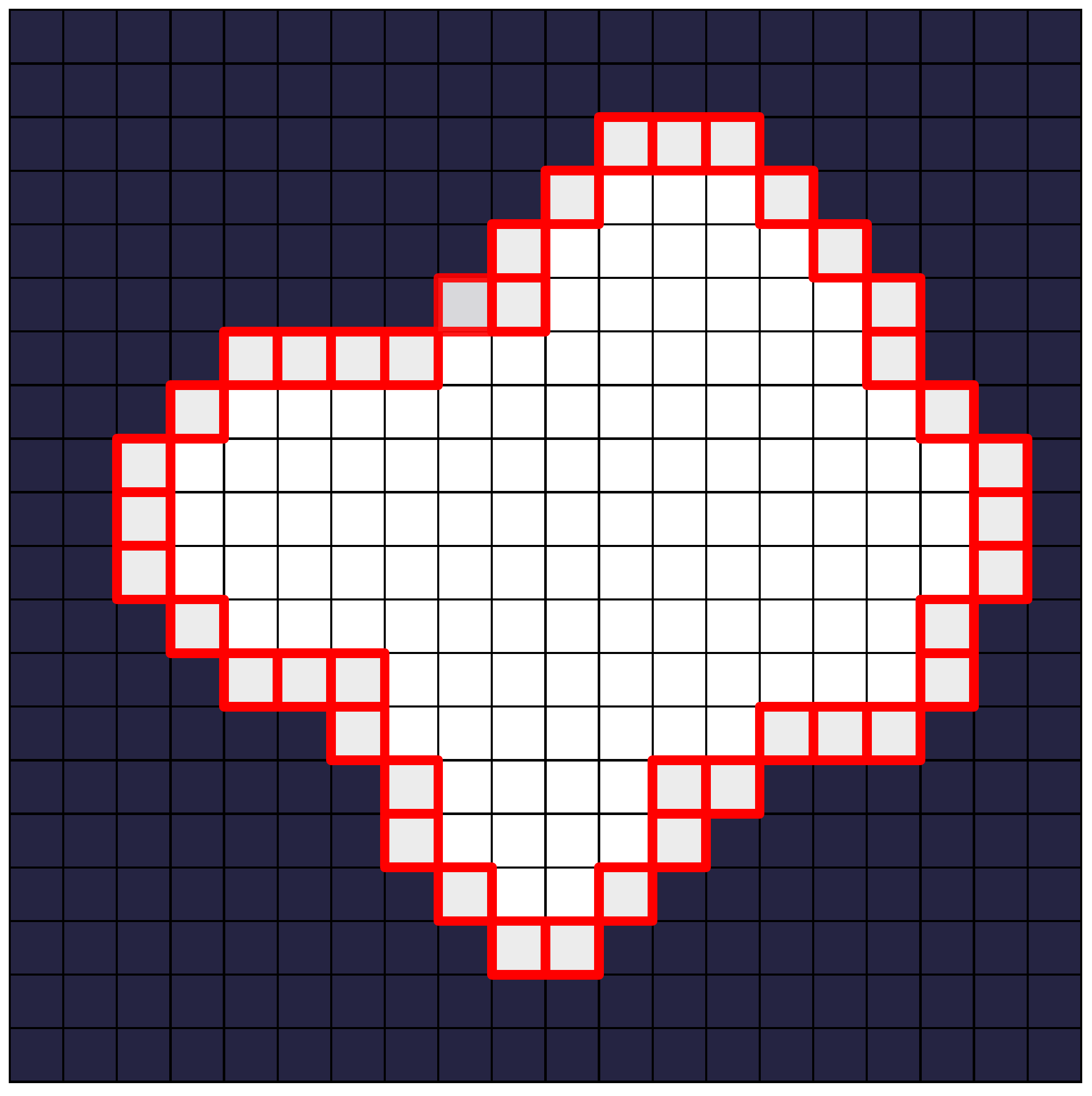} &
         \includegraphics[width=0.35\linewidth]{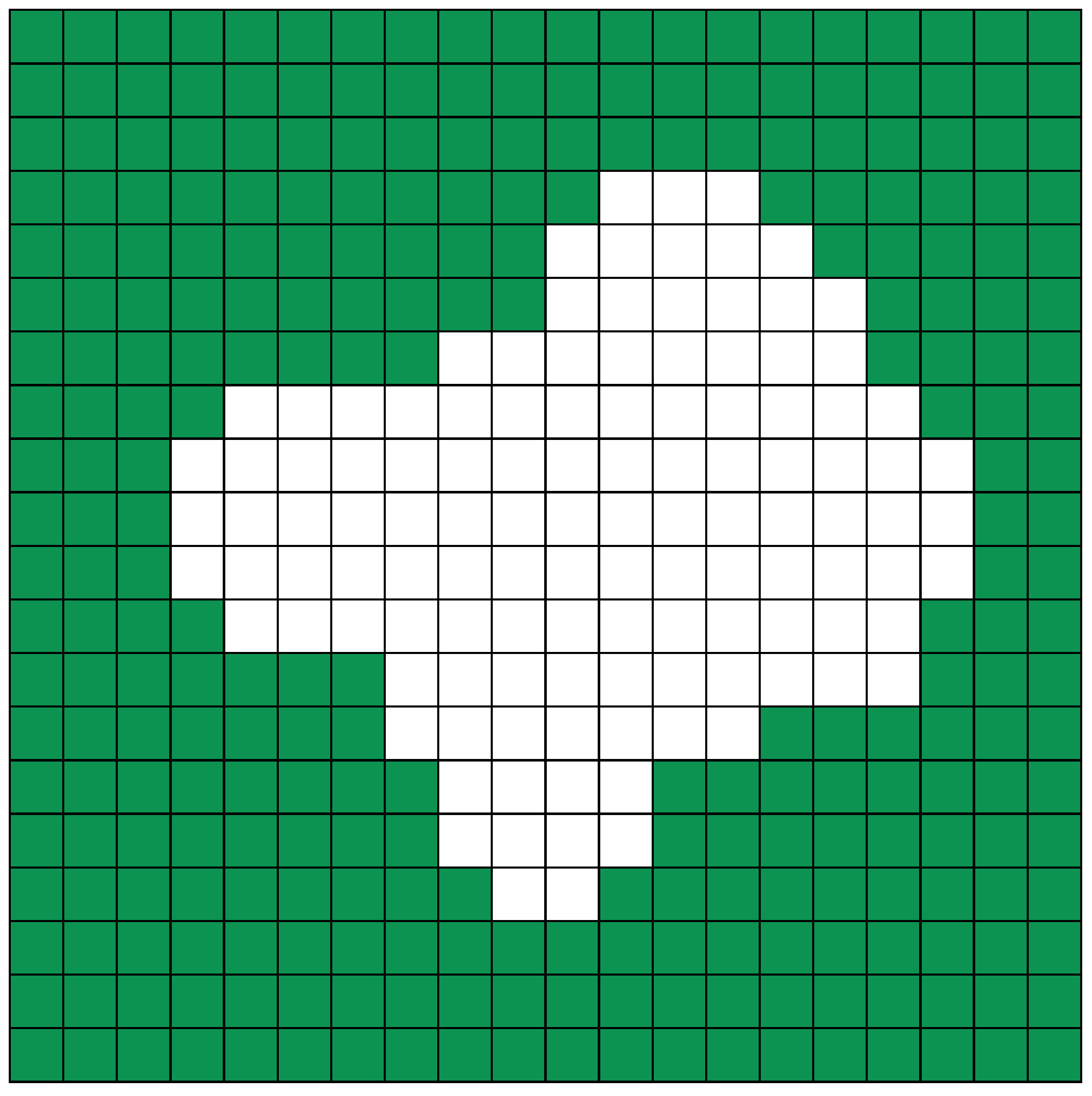}\\
         Mask Expansion & Content Feathering\\
    \end{tabular}
    \caption{Blending techniques integrated into the network architecture. For a given mask (Top Left), mask feathering (Top Right) is applied before stylization occurs, while dynamic mask expansion (Bottom Left) dilates the mask during each partial convolution in the style network. Content feathering (Bottom Right) maps the unstylized content feature to the pixels around the mask during the decoder phase. 
    Each of these approaches separately and collectively minimize boundary artifacts by giving the style network more context when assigning color values to the boundary pixels, enhancing output quality along the border.
    }
    \label{fig:blending_examples}
\end{figure}

\noindent Visualizations of these different approaches are visualized in Fig.~\ref{fig:blending_examples}.
Collectively, these techniques significantly enhance the integration between stylized and non-stylized regions, improving overall visual coherence and reducing border artifacts.
Additional implementation details for the blending techniques and are provided in the supplemental material.


\subsection{Multi-mask Style Transfer}\label{sec:multistyle}

Due to the unique nature of the partial convolution network, it is possible to provide multiple masks and multiple styles to the network in one pass.
Multi-mask style transfer enables the simultaneous application of distinct artistic styles to multiple, individually defined regions of an image.
By pairing each segmentation mask with a specific style, the method processes multiple masked regions in parallel using partial convolution layers.
Stylized features corresponding to each mask are merged at the feature level, with overlapping regions gracefully blended through weighted summation based on mask values, rather than being overwritten in a sequential order.
The final image is reconstructed via a single decoder conditioned on the merged feature representations, ensuring coherent integration of multiple styles within a single output.
This multi-mask network architecture is illustrated in the supplemental material.


\section{Results}\label{sec:Results}

\subsection{Qualitative Comparisons}


\newcommand{\imwidth}{1.05in}
\newcommand{\imheight}{0.725in}
\newcommand{\includegraphicsva}[1]{\raisebox{-.5\height}{\includegraphics[width=\imwidth,height=\imheight]{#1}}}
\newcommand{\includegraphicsvb}[1]{\raisebox{-.55\height}{\includegraphics[width=\imwidth,height=\imheight]{#1}}}
\newcommand{\includegraphicsvc}[1]{\raisebox{-.65\height}{\includegraphics[width=\imwidth,height=\imheight]{#1}}}
\newcommand{\includegraphicsvd}[1]{\raisebox{-.75\height}{\includegraphics[width=\imwidth,height=\imheight]{#1}}}
\newcommand{\includegraphicssq}[1]{\raisebox{-.55\height}{\includegraphics[width=\imheight,height=\imheight]{#1}}}


\begin{figure*}[t]
\begin{center}
\begin{tblr}{
  colspec={Q[c,1.3cm] *{8}{c}},
}
Original &
{\includegraphicsva{figures//results//Qualitative_vs_Quantitative_Comparisons/sa_224426.JPG}} &
\includegraphicsva{figures//results//Qualitative_vs_Quantitative_Comparisons/sa_223830.JPG} &
\includegraphicsva{figures//results//Qualitative_vs_Quantitative_Comparisons/sa_223982.JPG} &
\includegraphicsva{figures//results//Qualitative_vs_Quantitative_Comparisons/sa_223869.JPG} &
\includegraphicsva{figures//results//Qualitative_vs_Quantitative_Comparisons/sa_223776.JPG} 
\\

Mask and Style &
\includegraphicsvc{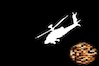} &
\includegraphicsvc{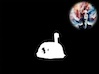} &
\includegraphicsvc{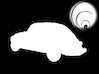}&
\includegraphicsvc{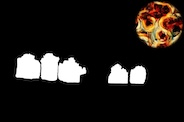} &
\includegraphicsvc{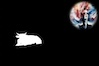} 
\\

Style-then-Mask \cite{li2017universal} &
\includegraphicsvd{figures//results//More_Styled_Images/helicopter7_style_then_mask.JPG} &
\includegraphicsvd{figures//results//Qualitative_vs_Quantitative_Comparisons/boat_style_then_mask.JPG} &
\includegraphicsvd{figures//results//More_Styled_Images/wv1_style_then_mask.JPG} &
\includegraphicsvd{figures//results//Qualitative_vs_Quantitative_Comparisons/trees_style_then_mask.JPG} &
\includegraphicsvd{figures//results//Qualitative_vs_Quantitative_Comparisons/cow_style_then_mask.JPG} 
\\

Mask-then-Style \cite{li2017universal} &
\includegraphicsvd{figures//results//More_Styled_Images/helicopter7_mask_then_style.JPG} &
\includegraphicsvd{figures//results//Qualitative_vs_Quantitative_Comparisons/boat_mask_then_style.JPG} &
\includegraphicsvd{figures//results//More_Styled_Images/wv1_mask_then_style.JPG} &
\includegraphicsvd{figures//results//Qualitative_vs_Quantitative_Comparisons/trees_mask_then_style.JPG} &
\includegraphicsvd{figures//results//Qualitative_vs_Quantitative_Comparisons/cow_mask_then_style.JPG} 
\\

SAM Styler \cite{Psychogyios2023SAMStylerEV} &
\includegraphicsvc{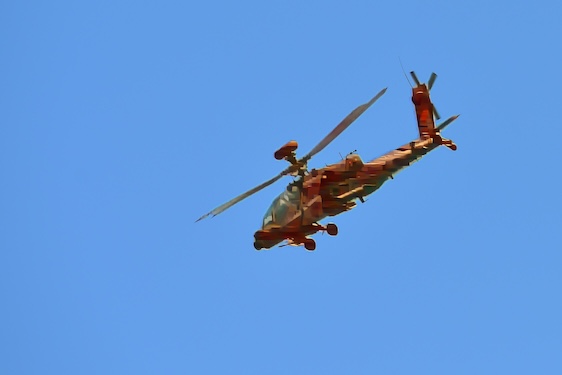} &
\includegraphicsvc{figures//results//Qualitative_vs_Quantitative_Comparisons/boat_sam_styler.JPG} &
\includegraphicsvc{figures//styleid//wv_samStyler.JPG} &
\includegraphicsvc{figures//results//Qualitative_vs_Quantitative_Comparisons/trees_sam_styler.JPG} &
\includegraphicsvc{figures//results//Qualitative_vs_Quantitative_Comparisons/cow_sam_styler.JPG} 
\\

StyTR$^2$ \cite{deng2022stytr2} &
\includegraphicssq{figures//styleid/helicopter_stytr2.JPG} &
\includegraphicssq{figures//results//Qualitative_vs_Quantitative_Comparisons/boat_stytr2.JPG} &
\includegraphicssq{figures//styleid//wv_stytr2.JPG} &
\includegraphicssq{figures//results//Qualitative_vs_Quantitative_Comparisons/trees_stytr2.JPG} &
\includegraphicssq{figures//results//Qualitative_vs_Quantitative_Comparisons/cow_stytr2.JPG} 
\\

StyleID \cite{styleID} & 
\includegraphicssq{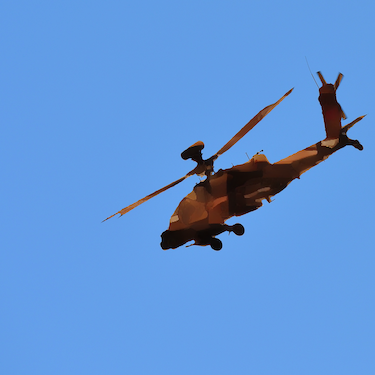} &
\includegraphicssq{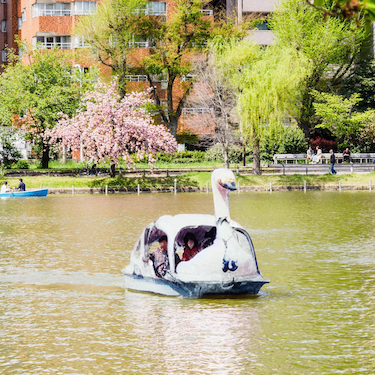}&
\includegraphicssq{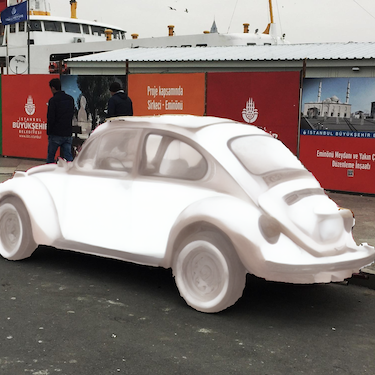} &
\includegraphicssq{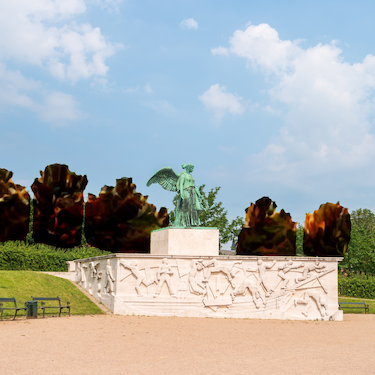} &
\includegraphicssq{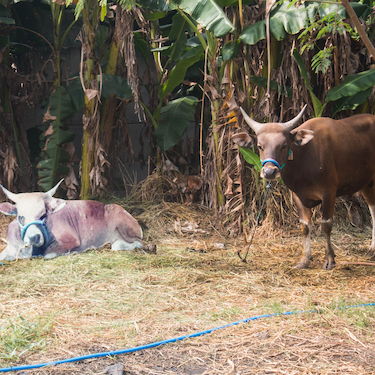} 
\\

Partial Conv
(Ours) &
\includegraphicsvc{figures//results//More_Styled_Images/helicopter7_partial_convolution.JPG} &
\includegraphicsvc{figures//results//Qualitative_vs_Quantitative_Comparisons/boat_partial_convolution.JPG} &
\includegraphicsvc{figures//results//More_Styled_Images/wv1_partial_convolution.JPG} &
\includegraphicsvc{figures//results//Qualitative_vs_Quantitative_Comparisons/trees_partial_convolution.JPG} &
\includegraphicsvc{figures//results//Qualitative_vs_Quantitative_Comparisons/cow_partial_convolution.JPG} \
\

\end{tblr}
\end{center}
\caption{Example outputs from different style techniques. Content images and style images give
the following stylized output for the masked region. Note the ability of our approach to consistently match style colors and features in the masked region of interest.}
\label{fig:results-grid}
\end{figure*}

Many visual comparisons between masked style algorithms are provided for different content, style, and
mask images in Fig. ~\ref{fig:results-grid}. In addition to style-then-mask and mask-then-style, we compare to masking two state-of-the-art style transfer methods: StyTr$^2$ \cite{deng2022stytr2}, a vision-transformer-based approach, and StyleID\cite{styleID} which uses a diffusion network.
We also compare to SAMstyler \cite{Psychogyios2023SAMStylerEV}, a recent method that is focused on masked stylization but relies on the slow optimization approach of Gatys \etal \cite{gatys2016image}. Additional qualitative examples are provided in the supplemental material.

From these examples, several observations can be made. First, our partial convolution approach consistently matches style features and color distributions well. Others method might match the style feature best in a given example but then perform poorly in other scenarios.
Second, as demonstrated in Section~\ref{sec:justification}, the various methods perform similarly when the region of interest is large or contains a color distribution similar to the background.



\subsection{Quantitative Comparisons}

\begin{figure}[t]
    \centering
    \begin{minipage}{1\columnwidth}
        
        \begin{minipage}[b]{0.23\columnwidth}
            \centering
            \includegraphics[width=\textwidth, height=1.5cm]{figures/introduction/style_then_mask_vs_partial_conv/1_style.jpg}
        \end{minipage}
        \begin{minipage}[b]{0.23\columnwidth}
            \centering
            \includegraphics[width=\textwidth, height=1.5cm]{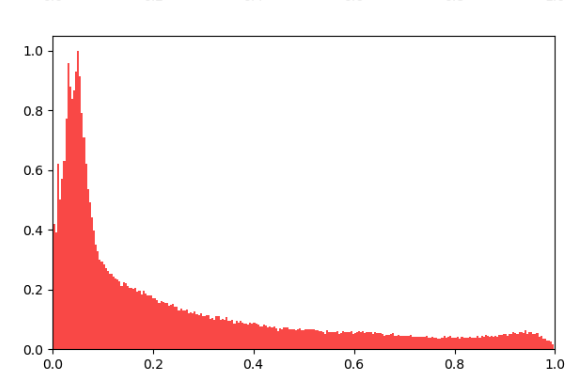}
        \end{minipage}
        \begin{minipage}[b]{0.23\columnwidth}
            \centering
            \includegraphics[width=\textwidth, height=1.5cm]{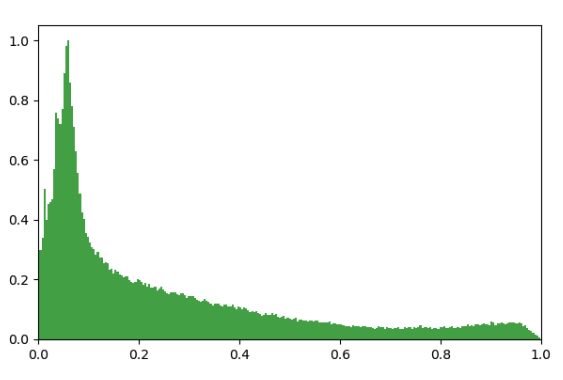}
        \end{minipage}
        \begin{minipage}[b]{0.23\columnwidth}
            \centering
            \includegraphics[width=\textwidth, height=1.5cm]{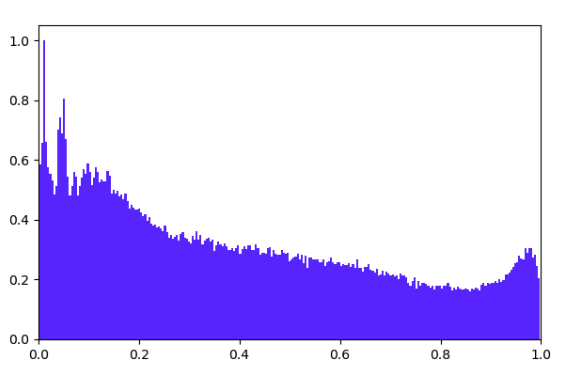}
        \end{minipage} \\[0.2em]

        \begin{minipage}[b]{0.23\columnwidth}
            \centering
            \includegraphics[width=\textwidth, height=1.5cm]{figures/introduction/style_then_mask_vs_partial_conv/1_style_then_mask.jpg}
        \end{minipage}
        \begin{minipage}[b]{0.23\columnwidth}
            \centering
            \includegraphics[width=\textwidth, height=1.5cm]{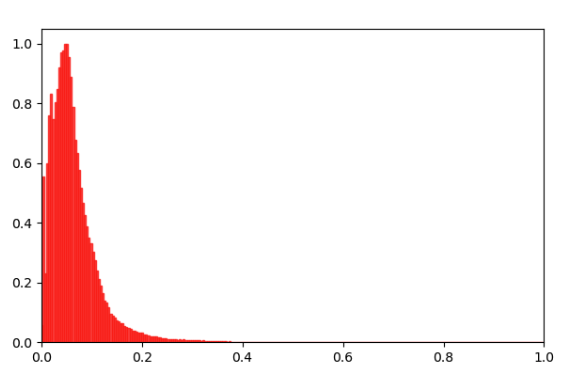}
        \end{minipage}
        \begin{minipage}[b]{0.23\columnwidth}
            \centering
            \includegraphics[width=\textwidth, height=1.5cm]{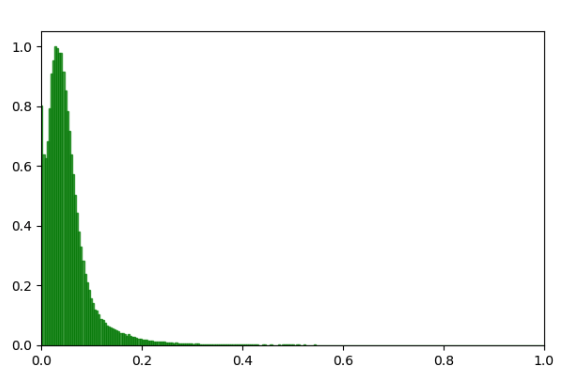}
        \end{minipage}
        \begin{minipage}[b]{0.23\columnwidth}
            \centering
            \includegraphics[width=\textwidth, height=1.5cm]{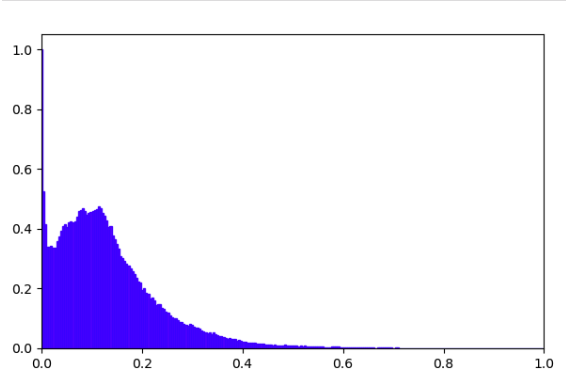}
        \end{minipage} 

        \begin{minipage}[t]{0.23\columnwidth}
            \centering
            \textbf{EMD}
        \end{minipage}
        \begin{minipage}[t]{0.23\columnwidth}
            \centering
            0.95
        \end{minipage}
        \begin{minipage}[t]{0.23\columnwidth}
            \centering
            0.96
        \end{minipage}
        \begin{minipage}[t]{0.23\columnwidth}
            \centering
            0.84
        \end{minipage} \\

        \begin{minipage}[b]{0.23\columnwidth}
            \centering
            \includegraphics[width=\textwidth, height=1.5cm]{figures/introduction/style_then_mask_vs_partial_conv/1_partial_conv.jpg}
        \end{minipage}
        \begin{minipage}[b]{0.23\columnwidth}
            \centering
            \includegraphics[width=\textwidth, height=1.5cm]{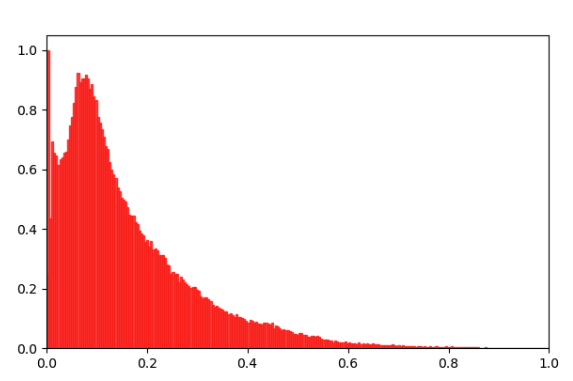}
        \end{minipage}
        \begin{minipage}[b]{0.23\columnwidth}
            \centering
            \includegraphics[width=\textwidth, height=1.5cm]{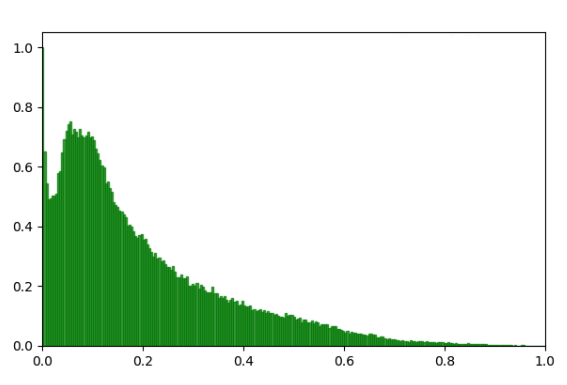}
        \end{minipage}
        \begin{minipage}[b]{0.23\columnwidth}
            \centering
            \includegraphics[width=\textwidth, height=1.5cm]{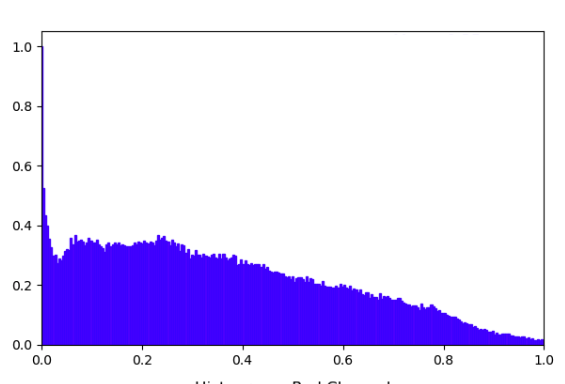}
        \end{minipage} 

        \begin{minipage}[t]{0.23\columnwidth}
            \centering
            \textbf{EMD}
        \end{minipage}
        \begin{minipage}[t]{0.23\columnwidth}
            \centering
            0.24
        \end{minipage}
        \begin{minipage}[t]{0.23\columnwidth}
            \centering
            0.31
        \end{minipage}
        \begin{minipage}[t]{0.23\columnwidth}
            \centering
            0.40
        \end{minipage} \\

        \begin{minipage}[t]{0.23\columnwidth}
            \centering
            \textbf{Image}
        \end{minipage}
        \begin{minipage}[t]{0.23\columnwidth}
            \centering
            \textbf{Red}
        \end{minipage}
        \begin{minipage}[t]{0.23\columnwidth}
            \centering
            \textbf{Green}
        \end{minipage}
        \begin{minipage}[t]{0.23\columnwidth}
            \centering
            \textbf{Blue}
        \end{minipage}

    \end{minipage}

    \caption[Style Region Histograms]{The style image color distributions (top) compared to the masked region color distributions for style-then-mask (middle) and partial convolution (bottom). The EMD distance for each channel suggests that partial convolution aligns more closely with the style image statistics. This is consistent with perceptual style loss metrics: style-then-mask = 729, partialconv = 622.}
    \label{fig:emd_histograms}
\end{figure}

Quantitative evaluations comparing Style-then-mask and PartialConv were conducted using Perceptual Style Loss metrics. Perceptual Style Loss measures how closely stylized images match intended styles as perceived by neural networks, where lower values reflect better style coherence \cite{johnson2016perceptual}. Earth Mover’s Distance (EMD) and sliced EMD can also be computed between the style image and the masked result, with lower EMD generally indicating a better stylization. Fig.~\ref{fig:emd_histograms} illustrates example histograms of the stylized regions for comparing the distributions with EMD. 

Comparisons were made across 500 images from the SA-1B \cite{kirillov2023sam} dataset with randomly selected masks. The average grayscale EMD, sliced EMD, and Perceptual Style Loss are presented in Tab.~\ref{tab:metric_results}. SamStyler \cite{Psychogyios2023SAMStylerEV} was excluded from the experiment because of its slow optimization time, as well as StyTr2 \cite{deng2022stytr2} and StyleID \cite{styleID} due to the restriction of only processing $512 \times 512$ inputs.
Results demonstrated partial convolution consistently achieved lower grayscale EMD, sliced EMD, and Perceptual Style Loss values, confirming its effectiveness in producing visually coherent and structurally consistent stylizations with reduced artifacts compared to traditional methods. 


\begin{table}[t]
    \centering
    \setlength{\tabcolsep}{2.9pt}
    \caption{Results on Masked Stylization averaged over a 500 image/mask dataset. Our partial convolution based approach out performs the standard approach on each metric.}
    \label{tab:metric_results}
    \begin{tabular}{lccc}
        \toprule
        Method & Gray EMD & Sliced EMD & Style Loss \\
        \midrule
        Style-then-mask & 0.121 & 0.168 & 760 \\
        PartialConv & \textbf{0.086} & \textbf{0.118} & \textbf{449} \\
        \bottomrule
    \end{tabular}
\end{table}

\subsection{Blending Results}

\begin{figure*}
    \centering
    \begin{minipage}[b]{0.24\textwidth}
        \centering
        \includegraphics[width=\textwidth,height=3.5cm]{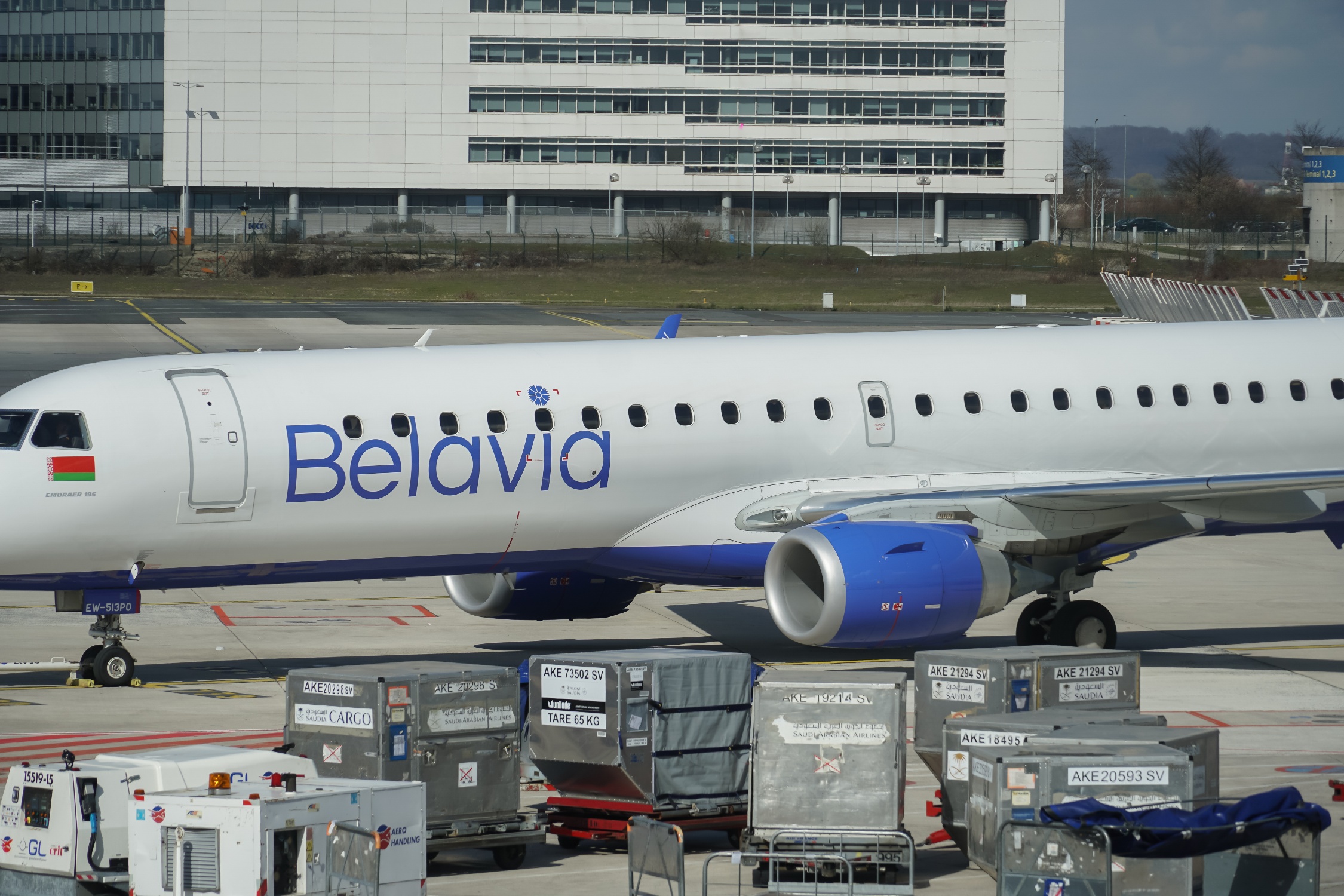}
    \end{minipage}
    \begin{minipage}[b]{0.24\textwidth}
        \centering
        \includegraphics[width=\textwidth,height=3.5cm]{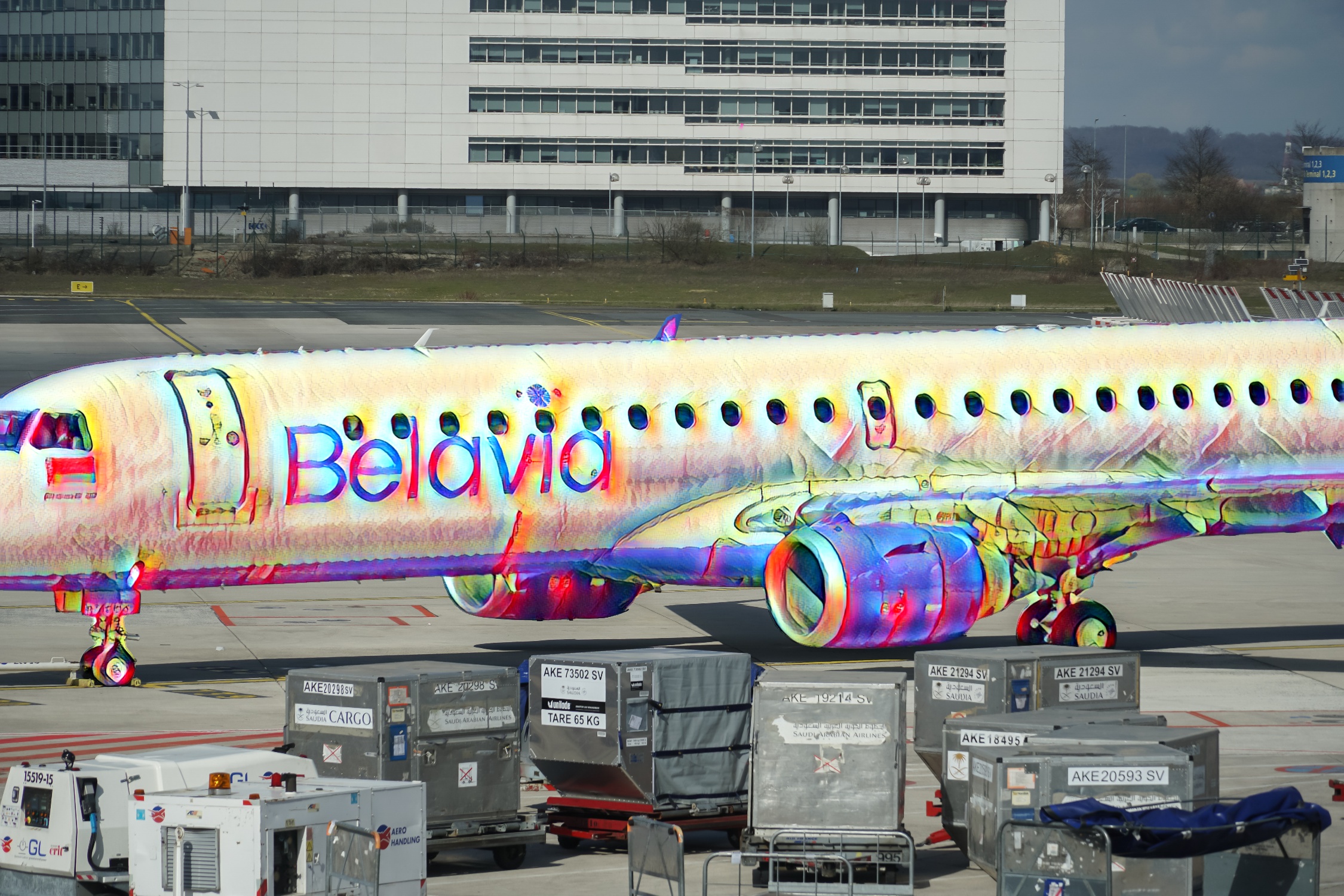}
    \end{minipage}
    \begin{minipage}[b]{0.24\textwidth}
        \centering
        \includegraphics[width=\textwidth,height=3.5cm]{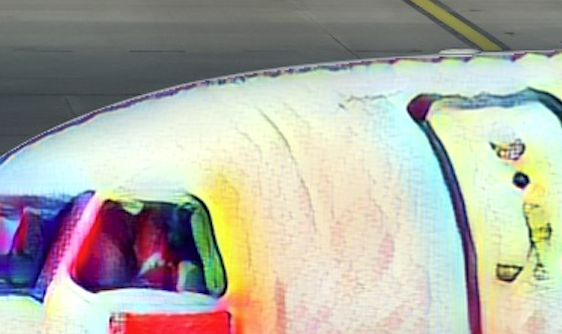}
    \end{minipage}
    \begin{minipage}[b]{0.24\textwidth}
        \centering
        \includegraphics[width=\textwidth,height=3.5cm]{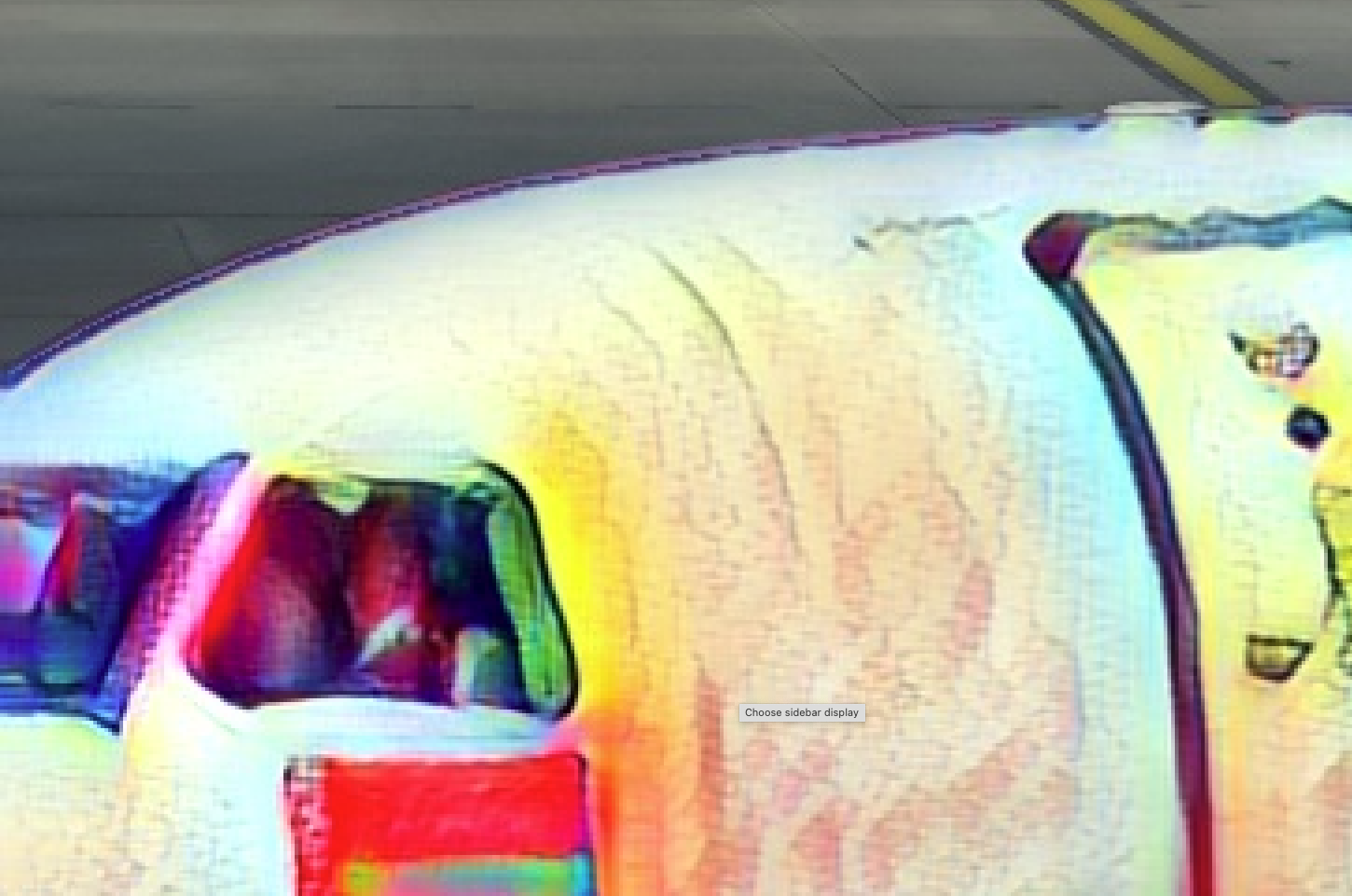}
    \end{minipage}
    \begin{minipage}[b]{0.24\textwidth}
        \centering
        \includegraphics[width=\textwidth,height=3.5cm]{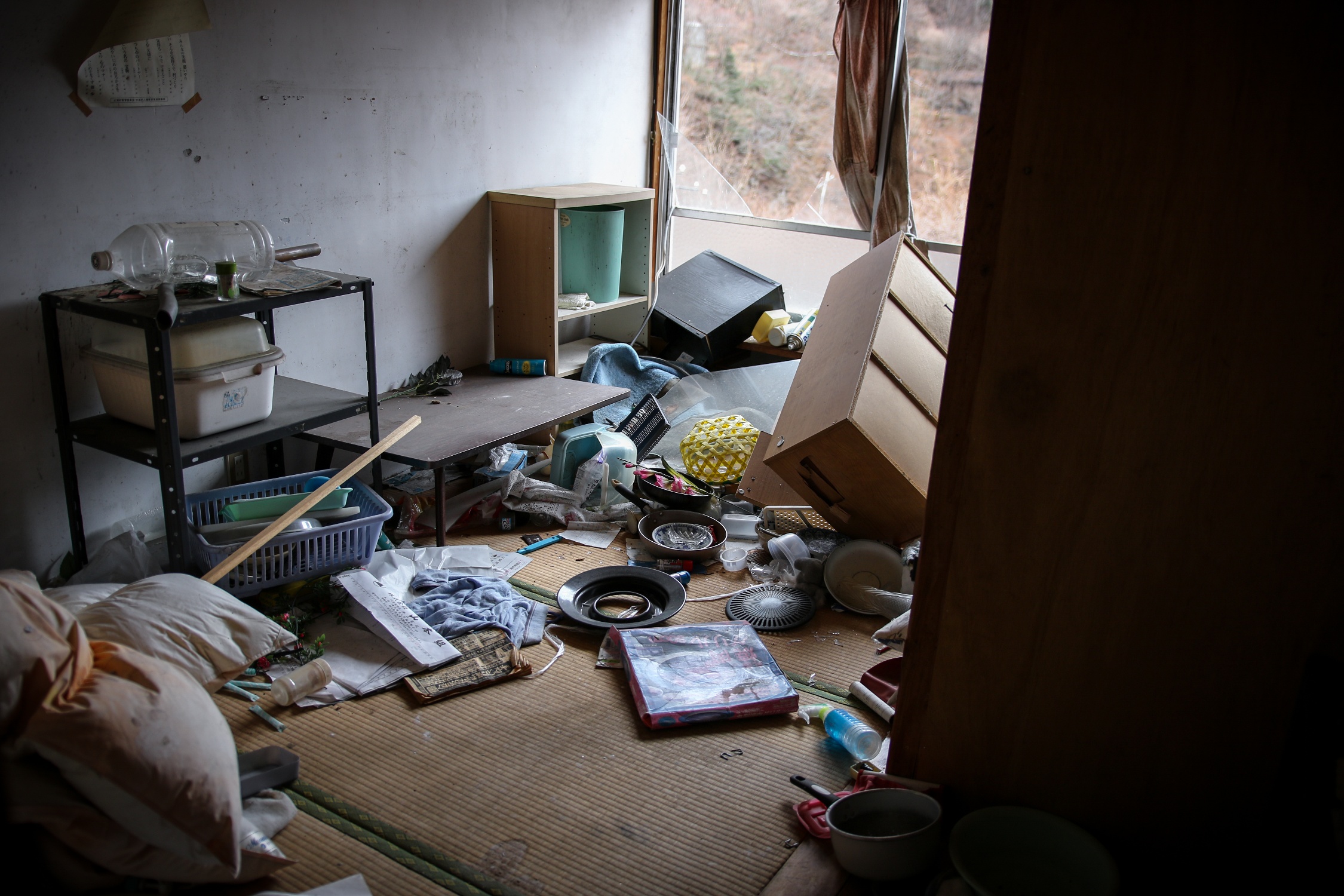}
    \end{minipage}
    \begin{minipage}[b]{0.24\textwidth}
        \centering
        \includegraphics[width=\textwidth,height=3.5cm]{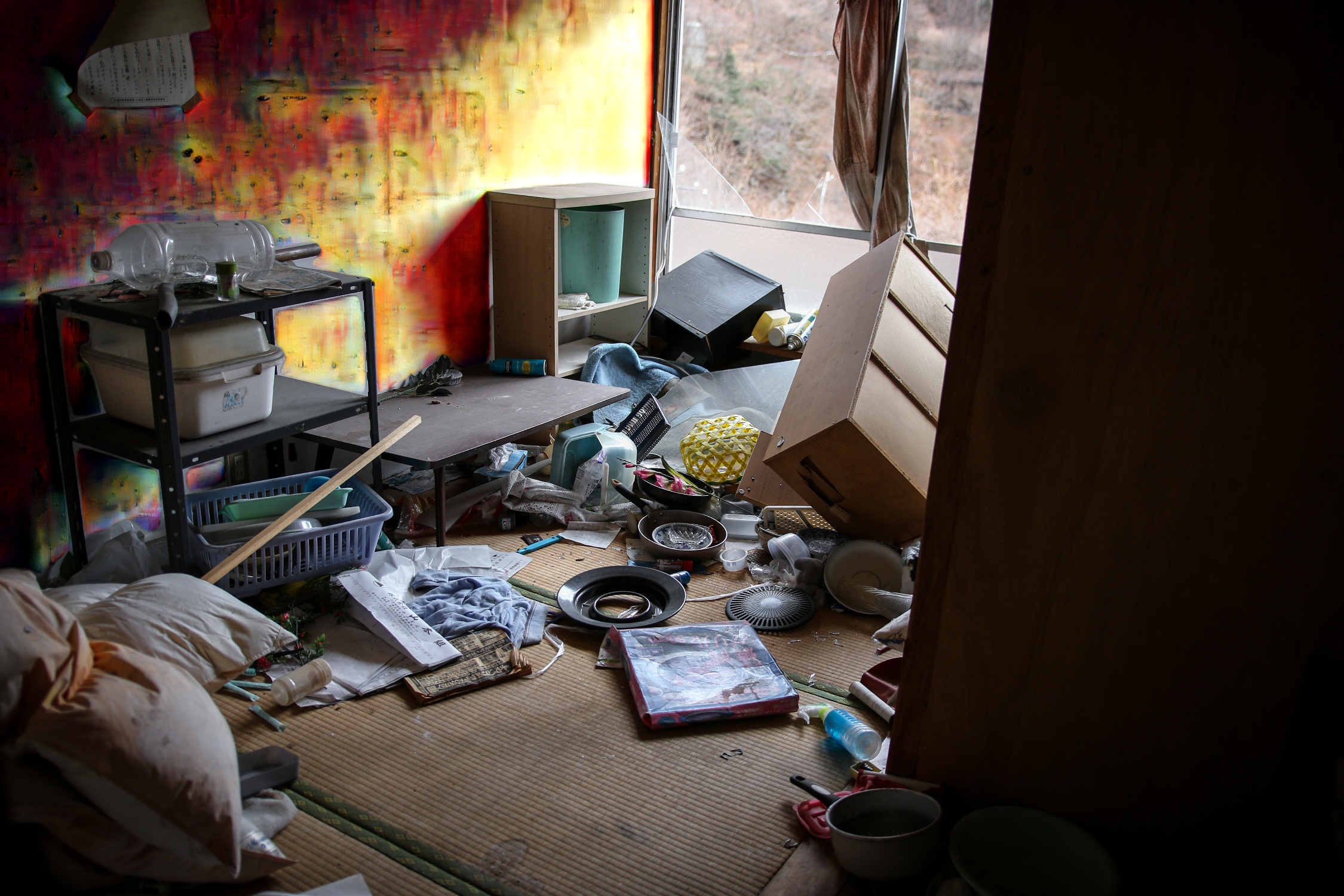}
    \end{minipage}
    \begin{minipage}[b]{0.24\textwidth}
        \centering
        \includegraphics[width=\textwidth,height=3.5cm]{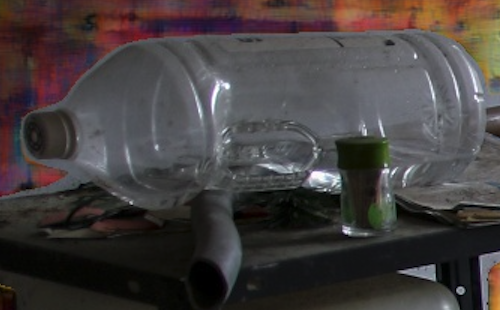}
    \end{minipage}
    \begin{minipage}[b]{0.24\textwidth}
        \centering
        \includegraphics[width=\textwidth,height=3.5cm]{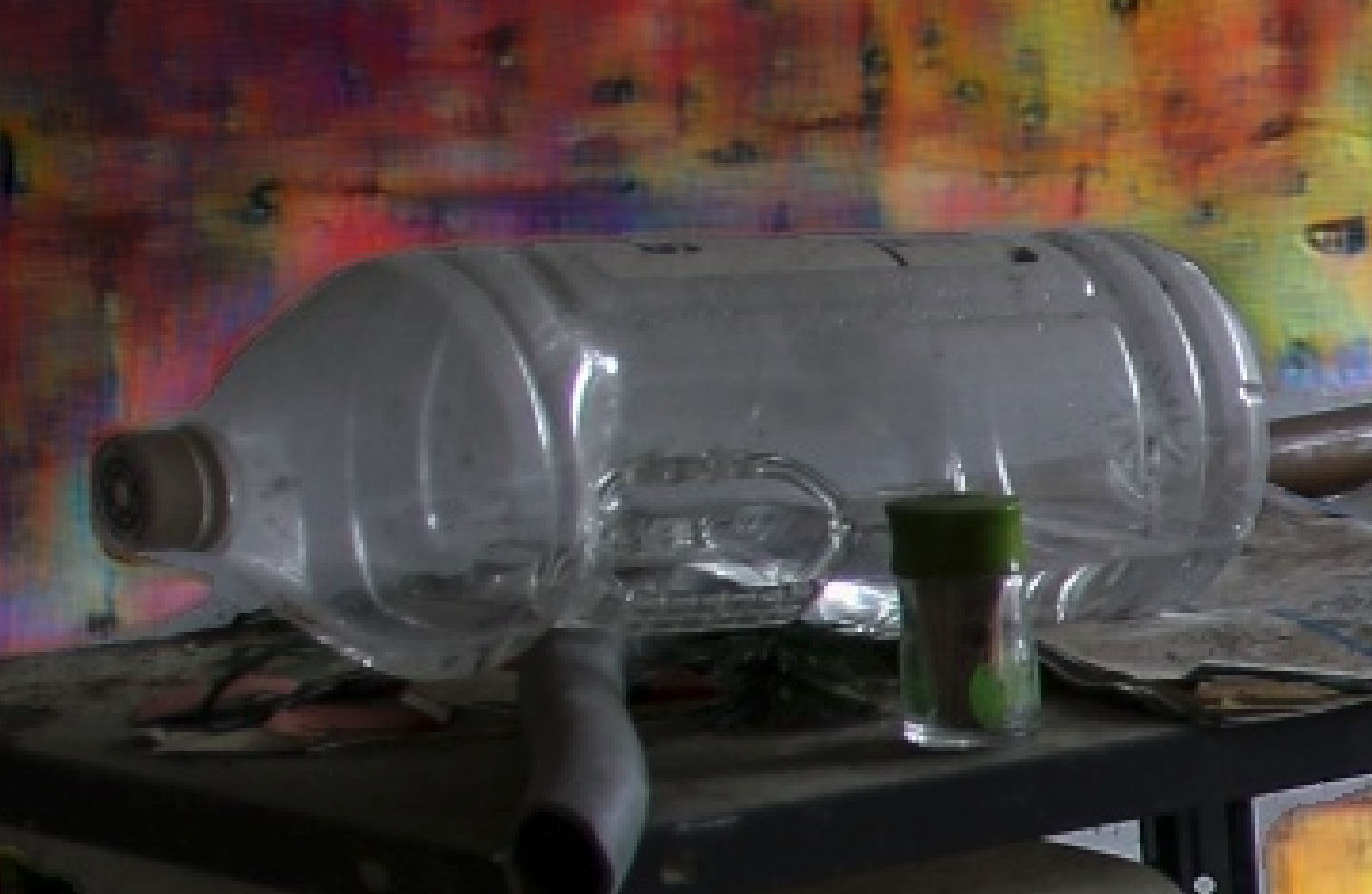}
    \end{minipage}
    \begin{minipage}[b]{0.24\textwidth}
        \centering
        \includegraphics[width=\textwidth,height=3.5cm]{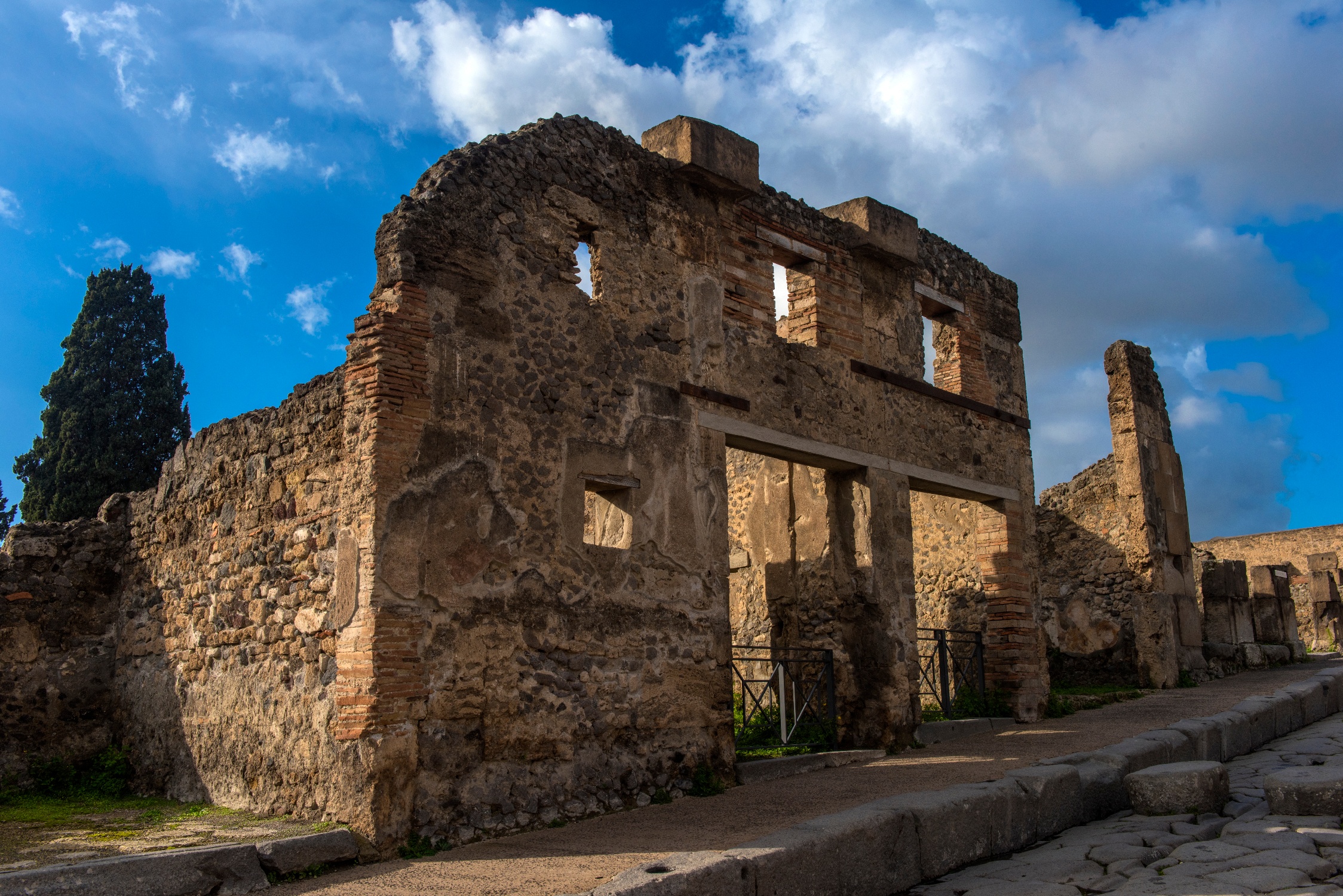}
    \end{minipage}
    \begin{minipage}[b]{0.24\textwidth}
        \centering
        \includegraphics[width=\textwidth,height=3.5cm]{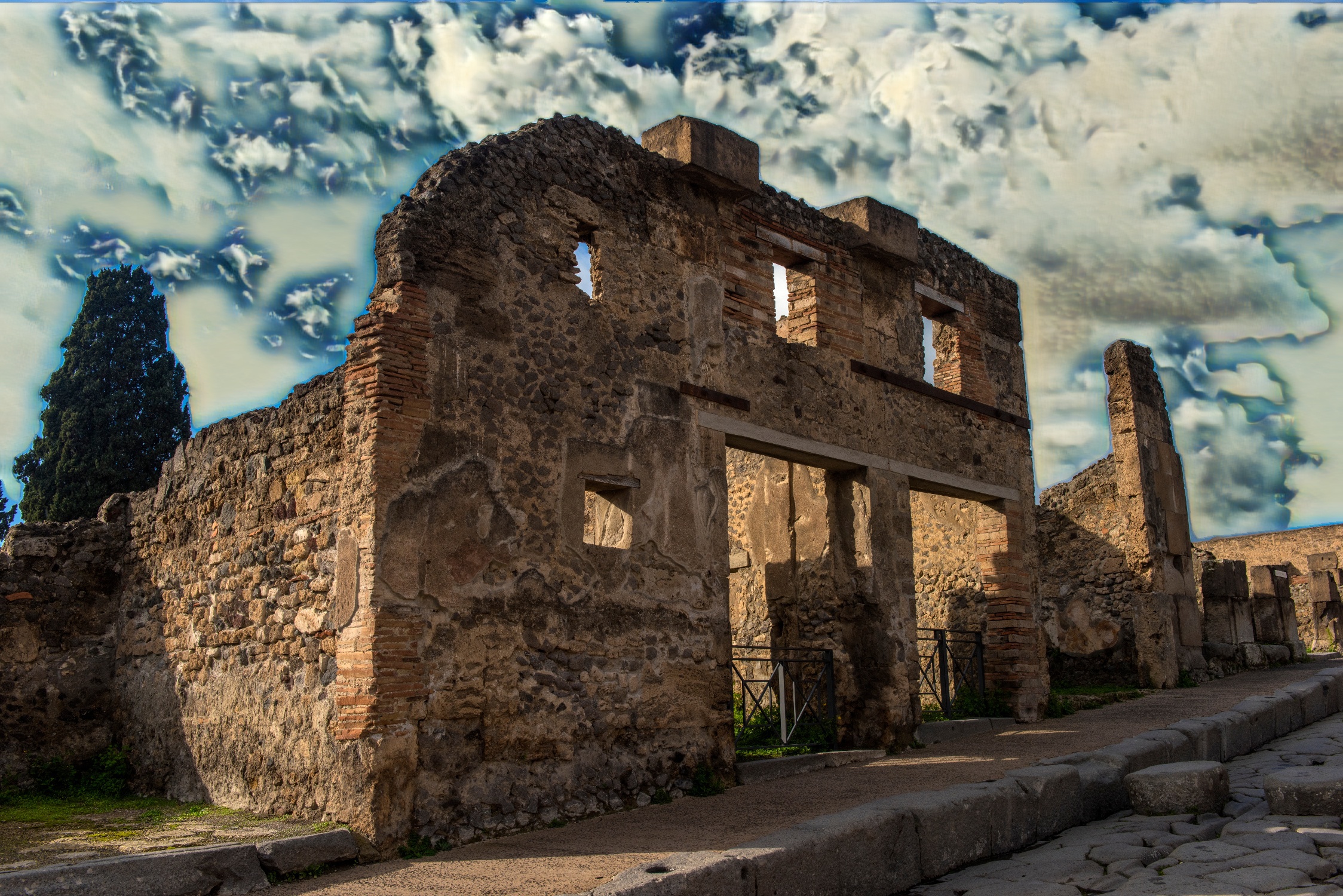}
    \end{minipage}
    \begin{minipage}[b]{0.24\textwidth}
        \centering
        \includegraphics[width=\textwidth,height=3.5cm]{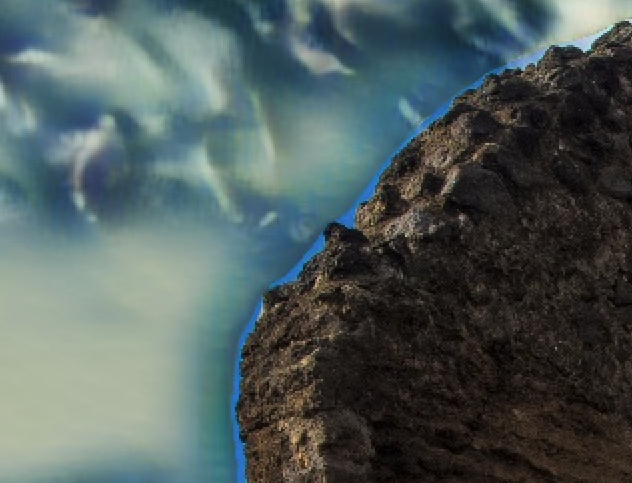}
    \end{minipage}
    \begin{minipage}[b]{0.24\textwidth}
        \centering
        \includegraphics[width=\textwidth,height=3.5cm]{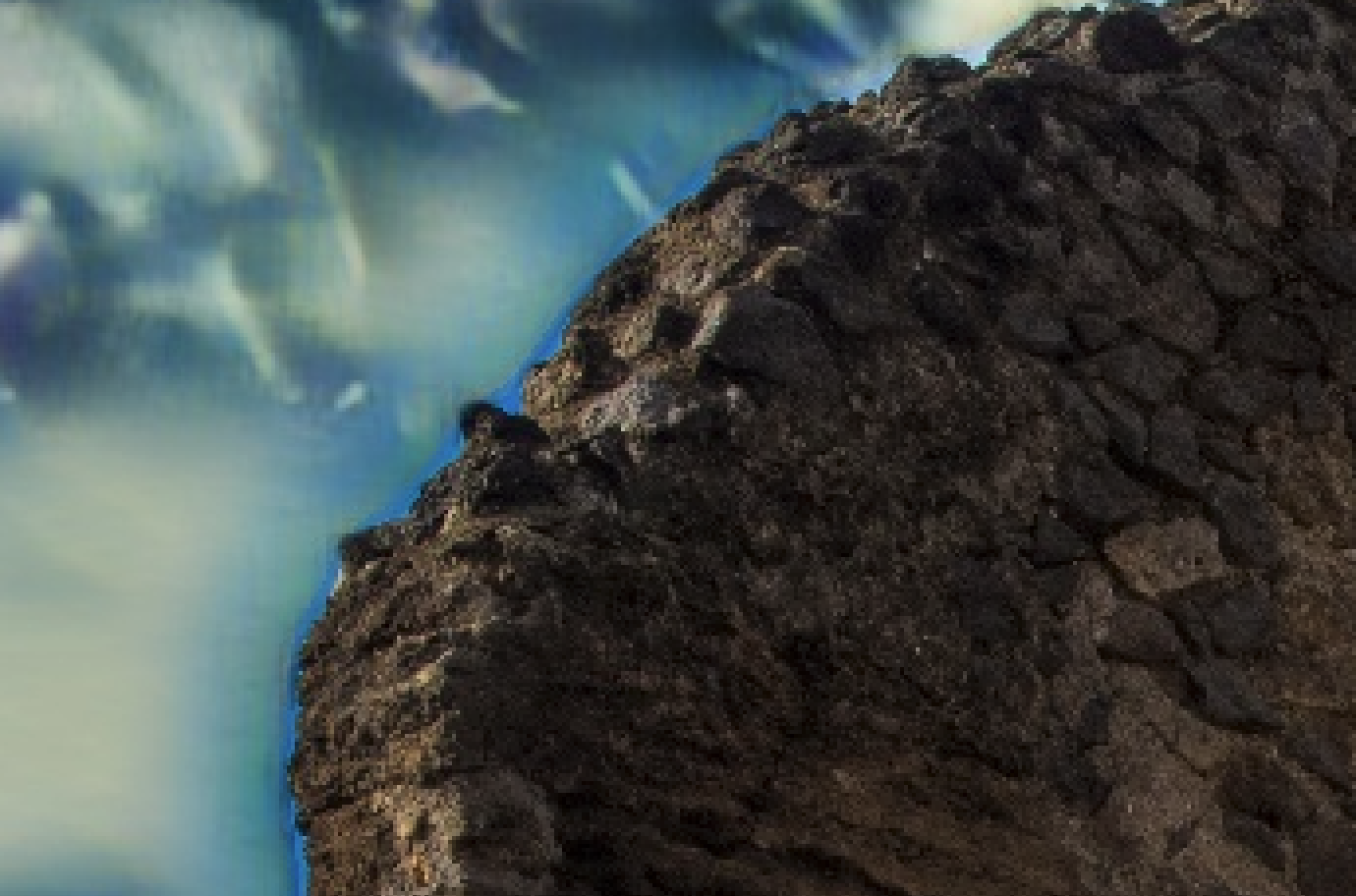}
    \end{minipage}
    \vspace{2mm}
    \begin{minipage}[t]{0.24\textwidth}
        \centering
        \text{\small(a) Content Image}
    \end{minipage}
    \begin{minipage}[t]{0.24\textwidth}
        \centering
        \parbox{\textwidth}{\centering (b) Stylized With Blending}
    \end{minipage}
    \begin{minipage}[t]{0.24\textwidth}
        \centering
        \text{\small(c) Without Techniques}
    \end{minipage}
        \begin{minipage}[t]{0.24\textwidth}
        \centering
        \text{\small(d) With Techniques}
    \end{minipage}
    \caption[Results of Three Stylization Combinations]{Visual results of the combination of all three blending techniques. Notice that the edges are smoother and border artifacts are reduced dramatically with all techniques applied.}
    \label{fig:all_combinations_results}
\end{figure*}

Fig.~\ref{fig:all_combinations_results} provides various examples of apply blending techniques to the style transfer method. Overall, the combined application of mask feathering before encoding, mask expansion during partial convolution, and content feathering led to most natural coherent results at the border of the styled area and original background content. 

To quantitatively evaluate the different blending techniques, the same 500 images from the SA-1B dataset were used. Different masks were randomly selected to guarantee a large border region. Each blending technique was considered in isolation and in combination with the others, leading to eight possible configurations in all. Tab.~\ref{tab:quantitative_results} provides the quantitative evaluations for each of the configurations. Two metrics are used to evaluate the border consistency. The first metric represents the average gradient magnitude across the RGB channels at the boundary between masked and unmasked regions. Lower values indicate smoother transitions in color and intensity along the border. The second metric is average color change in RGB along the border, with smaller values indicating smoother color transitions between the styled and unstyled areas. Based on the quantitative experiment, the content feathering in the decoder has the largest effect on decreasing border artifacts, while the combined approach of all three methods performed best overall.

\begin{table}[t]
\scriptsize
\setlength{\tabcolsep}{2.9pt}
\centering
\caption{Quantitative results for different feature combinations over the first 500 images of the SA-1B dataset. Incorporating all three blending techniques provides the lowest scores on the two boundary metrics.}
\begin{tabular}{l|cccccccc}
\toprule
\textbf{Feature} & \textbf{1} & \textbf{2} & \textbf{3} & \textbf{4} & \textbf{5} & \textbf{6} & \textbf{7} & \textbf{8} \\
\midrule
Feathering (Before)  & \ding{51} & \ding{55} & \ding{55} & \ding{51} & \ding{51} & \ding{55} & \ding{51} & \ding{55} \\
Expansion (During)   & \ding{55} & \ding{51} & \ding{55} & \ding{51} & \ding{55} & \ding{51} & \ding{51} & \ding{55} \\
Feathering (Decoder) & \ding{55} & \ding{55} & \ding{51} & \ding{55} & \ding{51} & \ding{51} & \ding{51} & \ding{55} \\
\midrule
Grad. Magn.          & 142.9 & 135.3 & 90.38  & 129.9 & 88.03  & 84.61  & \textbf{82.65}  & 157.1 \\
Color Cont.          & 28.84  & 28.02  & 33.50  & 27.01  & 30.93  & 27.98  & \textbf{26.23}  & 31.19 \\
\bottomrule
\end{tabular}
\label{tab:quantitative_results}
\end{table}







\subsection{Multi-mask results}

\begin{figure*}
    \centering
    \begin{tabular}{cc}
    \includegraphics[width=0.45\linewidth]{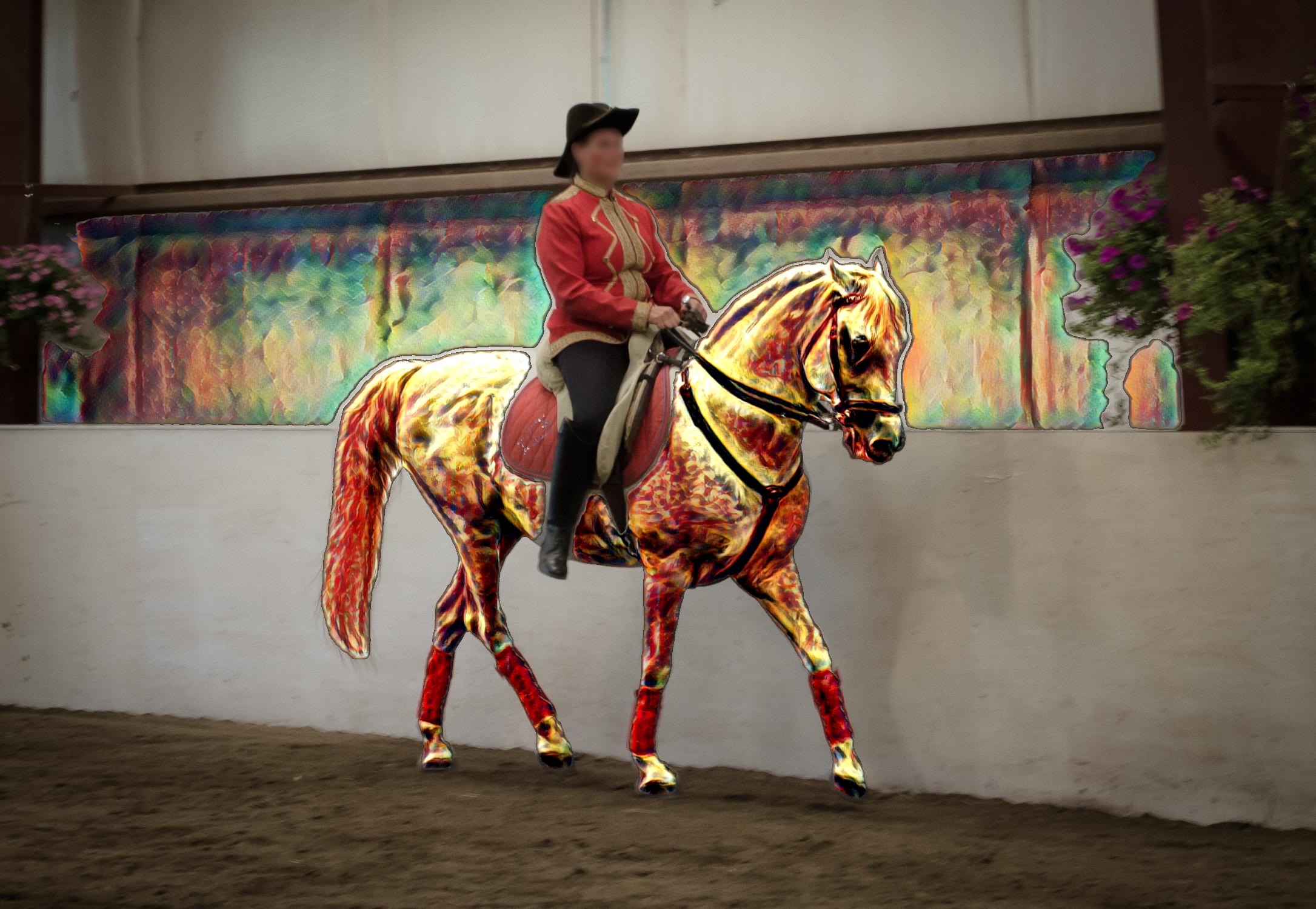}
    &
     \includegraphics[width=0.45\linewidth]{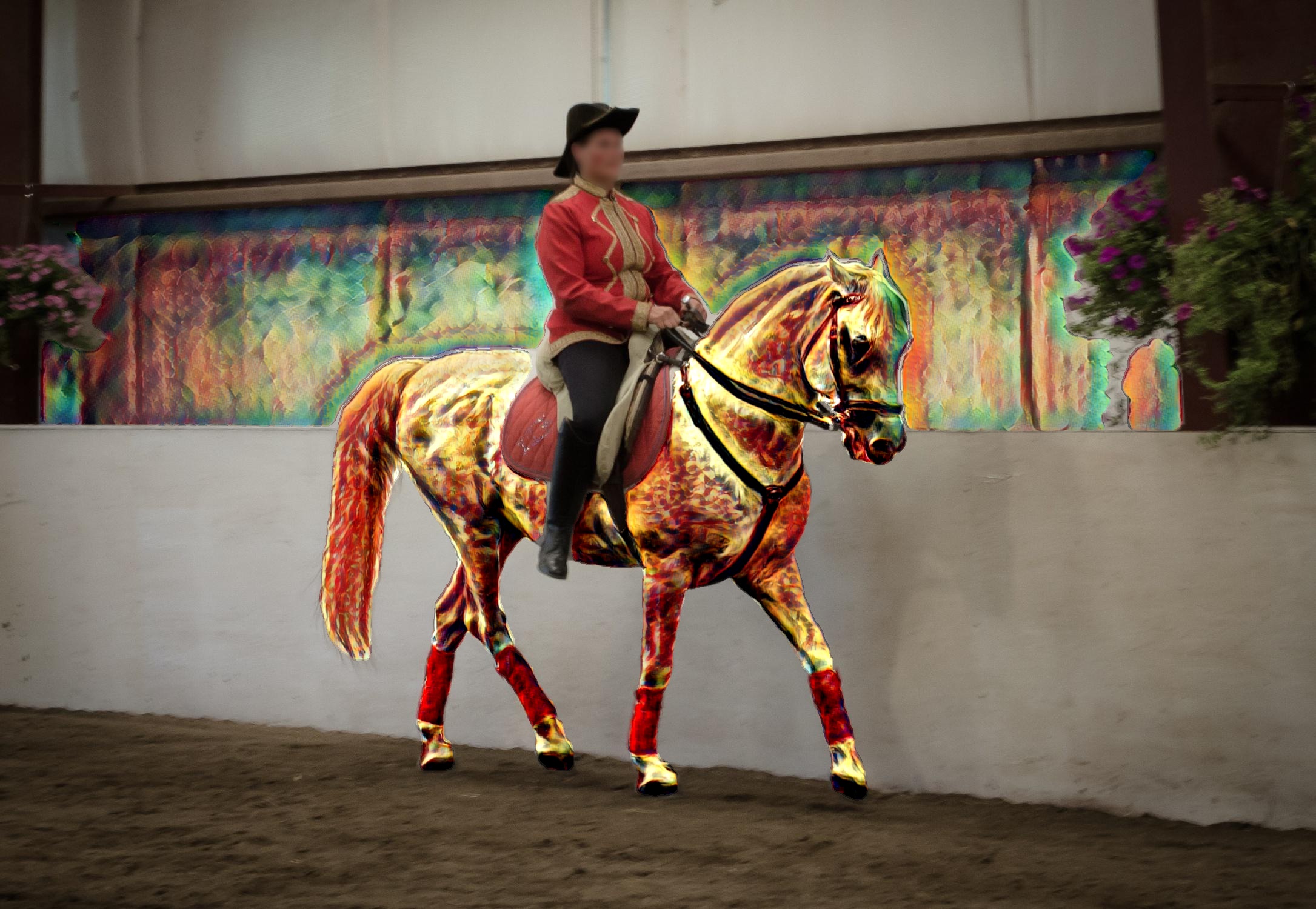} 
     \\
     Sequential Styling & Parallel Styling
    \end{tabular}
    \caption{Visual results of sequential styling compared to our parallel multi-mask/multi-style approach. Different stylized regions are blended around the touching borders of the masks which reduces stylization interruptions and avoids boundary artifacts. 
    }
    \label{fig:seq_vs_multimask}
\end{figure*}

As outlined in Section~\ref{sec:multistyle}, our approach allows for stylizing two regions in parallel during the decoder stage, allowing for blendings between two styles with adjacent masked regions.
In the multi-mask and multi-style scenario, each image region receives a unique style simultaneously using partial convolution with mask feathering and expansion techniques. 
The example in Fig.~\ref{fig:seq_vs_multimask} demonstrates seamless integration of multiple styles, with smooth transitions and minimal border artifacts, highlighting the practical effectiveness of this combined method in complex stylization tasks.


\section{Conclusion}

In this work, we have presented a new stylization method that is specifically designed for masked regions of an image. We have shown that it stylizes better than other approaches in the masked region by more closely matching style image statistics. We have demonstrated various blending techniques and other practical improvements. This can lead to improved user experiences and work flows when working with style transfer algorithms in standard image applications. Future work could continue to develop mask-focused style transfer algorithms while incorporating modern network designs such as vision transformers and diffusion models.






{
    \small
    \bibliographystyle{ieeenat_fullname}
    \bibliography{main}
}

\end{document}


\title{Improving Masked Style Transfer using Blended Partial Convolution - Supplemental Material}

\maketitle

\section{Implementing Blending Techniques}

While applying style transfer to segmented regions using partial convolution improves overall visual quality, artifacts can still emerge along the border of the regions. These artifacts are visual discontinuities that occur due to abrupt changes between the stylized (masked) and non-stylized (unmasked) areas. The blending techniques described in the paper mitigate these artifacts. 

\begin{figure}[h]
    \centering
    \includegraphics[width=0.95\linewidth]{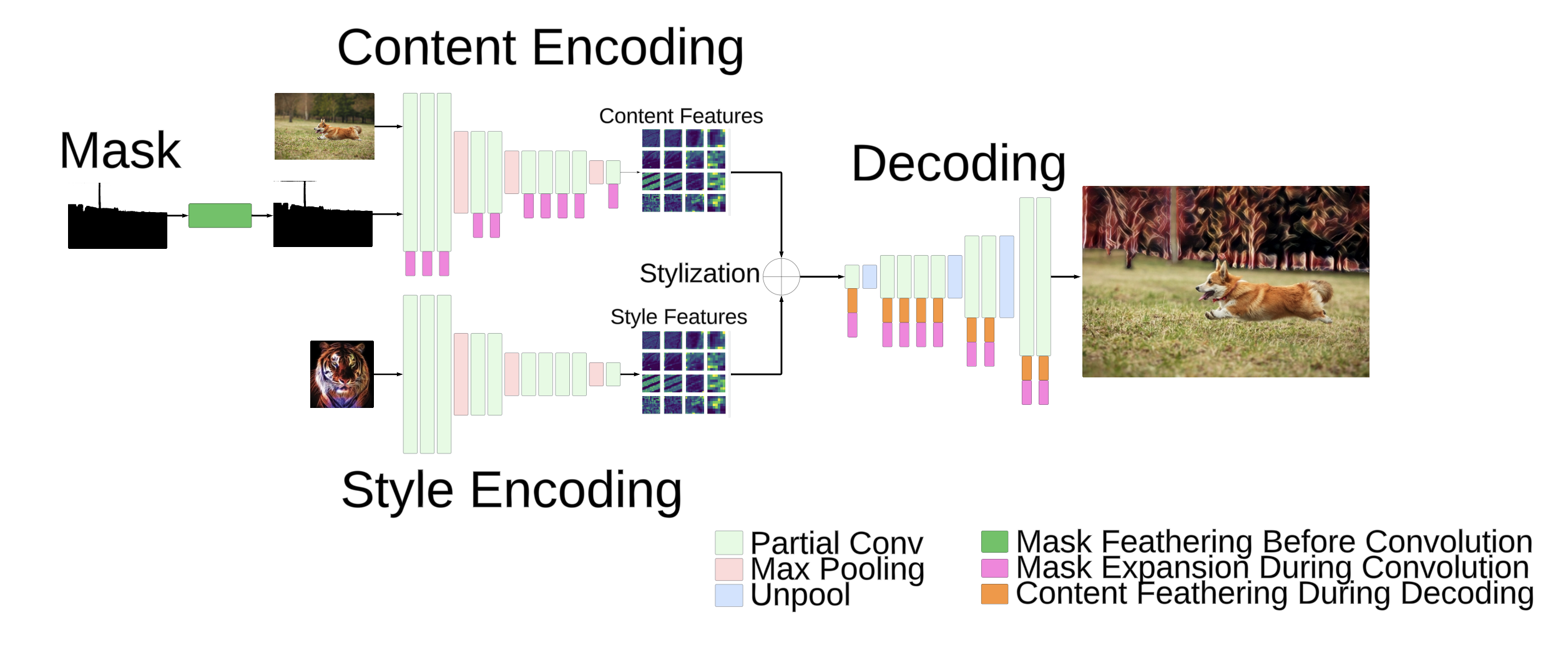}
    \caption[Blending Strategy Integration]{Blending techniques integrated into the network architecture. Mask feathering is applied before encoding, while dynamic mask expansion occurs during partial convolution. Content feathering occurs during the decoding stage to provide output context. Collectively these techniques minimize boundary artifacts and enhancing output quality.} 
    \label{fig:feathering_network}
\end{figure}

Each blending technique involves feathering or expanding the mask in some way. These techniques, however, are applied at different stages of the convolution network. \textbf{Mask Feathering} is just applied to the initial mask before the style network is applied. \textbf{Mask Expansion} occurs at each layer of the network. \textbf{Content Feathering} occurs only in the decoder stage. This is illustrated in the Fig.~\ref{fig:feathering_network}.

\textbf{Mask Feathering} is a simple operation implemented using OpenCV's morphology package. We used a feathering kernel with a size of 5 pixels.

\textbf{Mask Expansion} is implemented in PyTorch within the partial convolution operation. A max pooling operation with a kernel size of 3 and a stride of 1 is used to expand the mask (similar to a dilation in OpenCV). That mask is then used during the convolution, but not stored during the following layers, essentially updating the feature information 1 pixel around the original mask at each layer.

\textbf{Content Feathering} is implemented in PyTorch within the decoder stage. Since the style network is autoencoder based, the content image can be fed in parallel to the autoencoder without performing the stylization process. These unstylized content features can be mapped around the stylized masked features using simple copy operations during the decoder stage.

\section{Implementing Multi-mask Style Transfer}

\begin{figure}[h]
    \centering
    \includegraphics[width=0.95\linewidth]{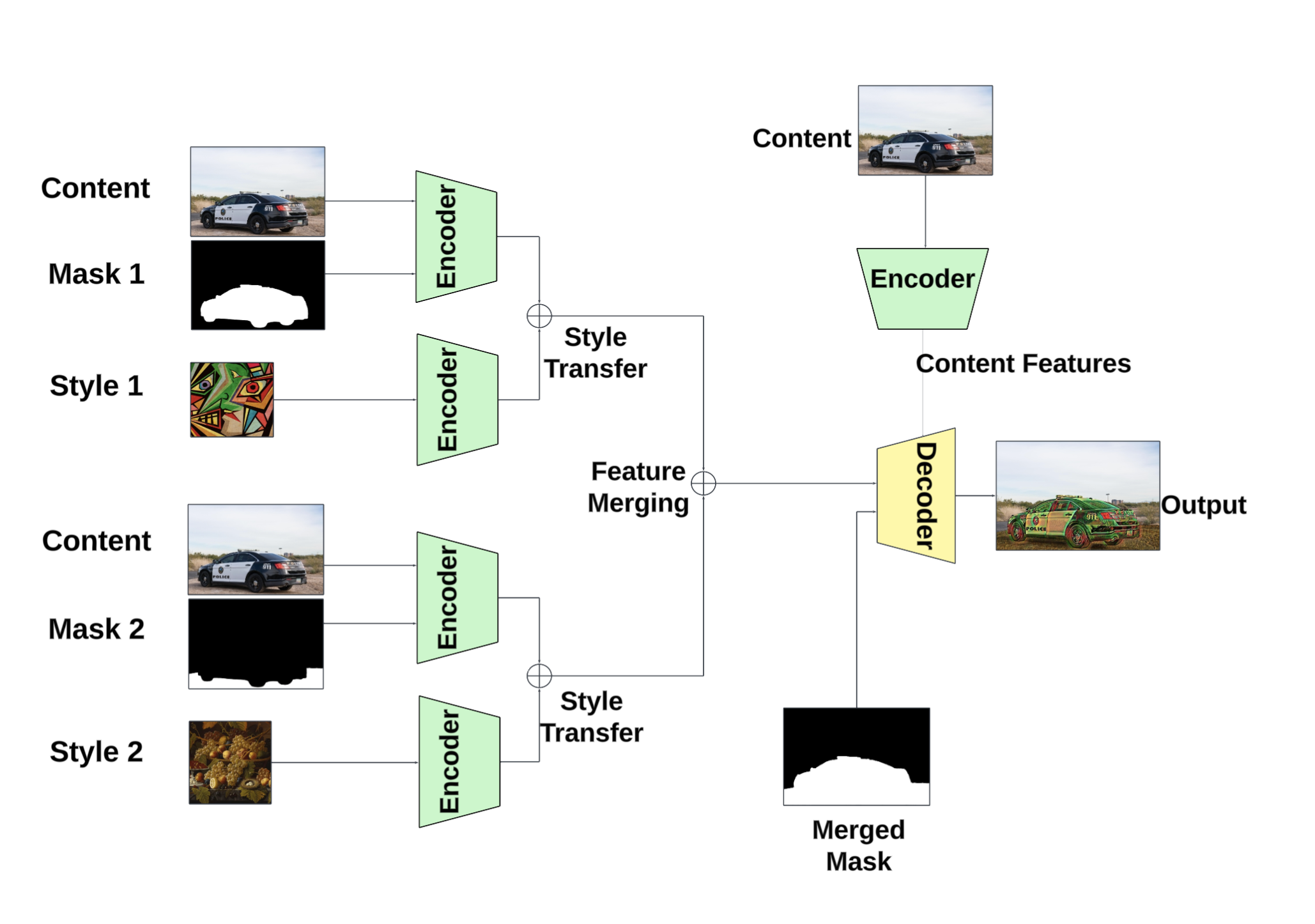}
    \caption[Multi-Mask and Multi-Style Framework]{Multi-mask and Multi-style parallel processing framework. Different regions of the content image are associated with distinct style images, processed in parallel, and merged at the feature level before decoding into a final stylized image.}
    \label{fig:multimask_multistyle_architecture}
\end{figure}

Our multi-mask and multi-style approach is simple to implement given the existing pieces of our framework. Each mask and style is fed through a separate encoder and stylizer in parallel. The simple modification is to merge the separate masked features together before sending them through a single decoder. For improved results, content features are also mapped into the decoder for the previously described blending technique. This whole process is visualized in Fig.~\ref{fig:multimask_multistyle_architecture}.

\section{Additional Style Transfer Results}

\begin{figure*}[t]
\begin{center}
       \begin{tabular}{cccccc}

        \\
        \includegraphics[width=0.15\linewidth, height=0.7in]{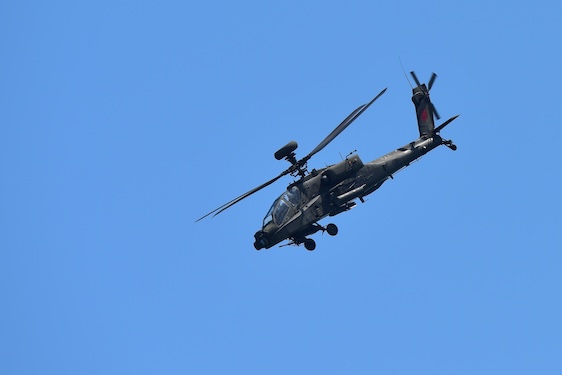}
        &
        \includegraphics[width=0.15\linewidth, height=0.7in]{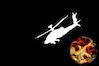}
        &
        \includegraphics[width=0.15\linewidth, height=0.7in]{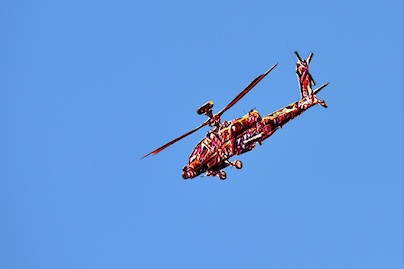}
        &
        \includegraphics[width=0.15\linewidth, height=0.7in]{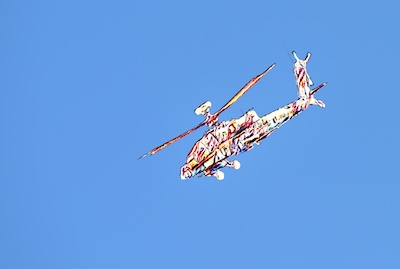}
        &
        \includegraphics[width=0.7in, height=0.7in]{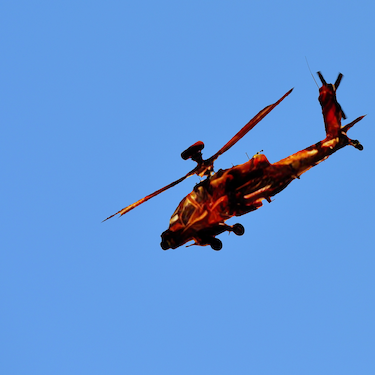}
        &
        \includegraphics[width=0.15\linewidth, height=0.7in]{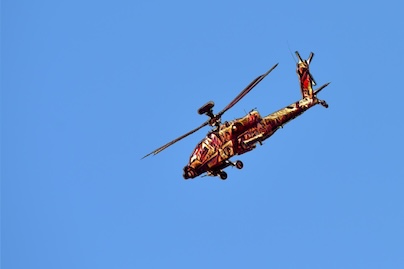}
        \\
        \includegraphics[width=0.15\linewidth, height=0.7in]{figures//results//More_Styled_Images/sa_224426.JPG}
        &
        \includegraphics[width=0.15\linewidth, height=0.7in]{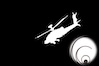}
        &
        \includegraphics[width=0.15\linewidth, height=0.7in]{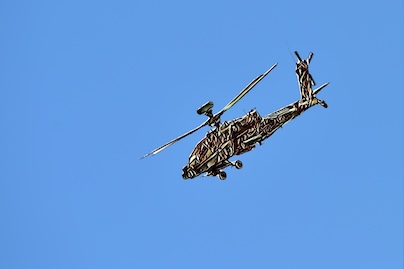}
        &
        \includegraphics[width=0.15\linewidth, height=0.7in]{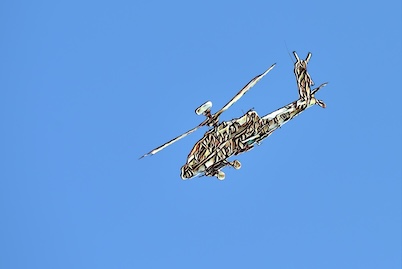}
        &
        \includegraphics[width=0.7in, height=0.7in]{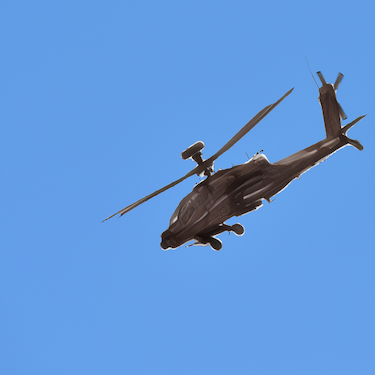}
        &
        \includegraphics[width=0.15\linewidth, height=0.7in]{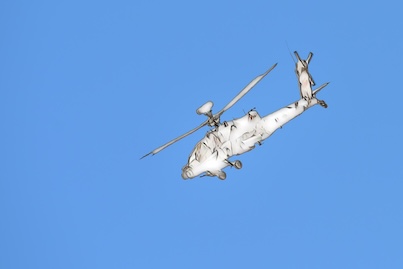}
        \\
        \includegraphics[width=0.15\linewidth, height=0.7in]{figures//results//More_Styled_Images/sa_224426.JPG}
        &
        \includegraphics[width=0.15\linewidth, height=0.7in]{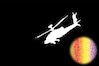}
        &
        \includegraphics[width=0.15\linewidth, height=0.7in]{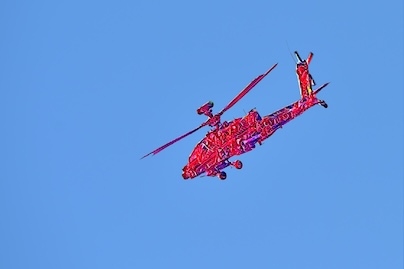}
        &
        \includegraphics[width=0.15\linewidth, height=0.7in]{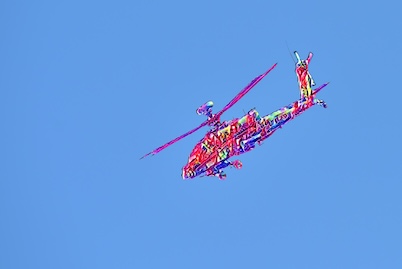}
        &
        \includegraphics[width=0.7in, height=0.7in]{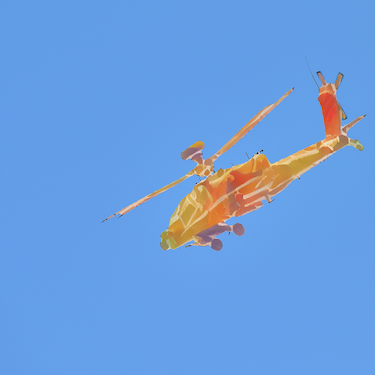}
        &
        \includegraphics[width=0.15\linewidth, height=0.7in]{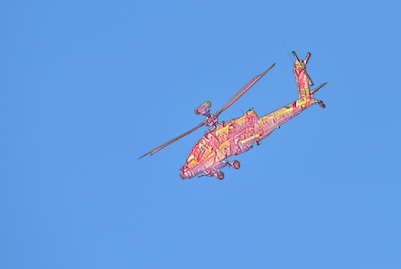}
        \\
        \includegraphics[width=0.15\linewidth, height=0.7in]{figures//results//More_Styled_Images/sa_224426.JPG}
        &
        \includegraphics[width=0.15\linewidth, height=0.7in]{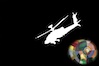}
        &
        \includegraphics[width=0.15\linewidth, height=0.7in]{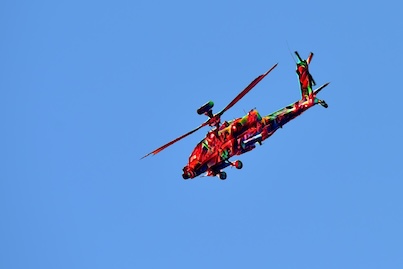}
        &
        \includegraphics[width=0.15\linewidth, height=0.7in]{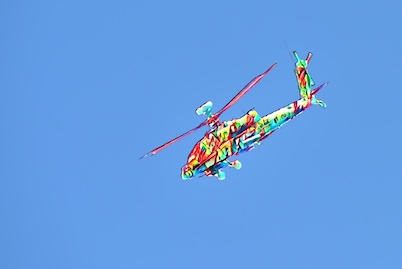}
        &
        \includegraphics[width=0.7in, height=0.7in]{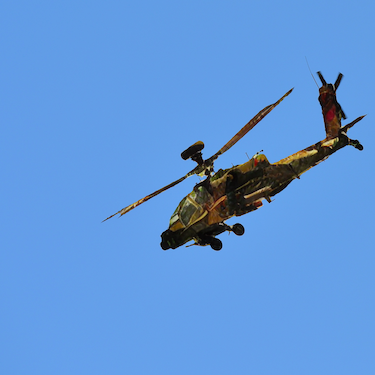}
        &
        \includegraphics[width=0.15\linewidth, height=0.7in]{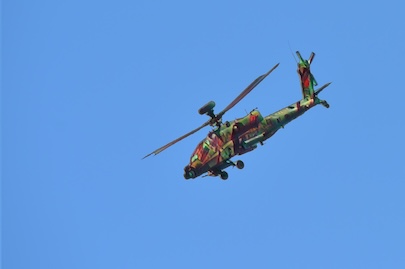}
        \\
        \includegraphics[width=0.15\linewidth, height=0.7in]{figures//results//More_Styled_Images/sa_224426.JPG}
        &
        \includegraphics[width=0.15\linewidth, height=0.7in]{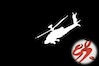}
        &
        \includegraphics[width=0.15\linewidth, height=0.7in]{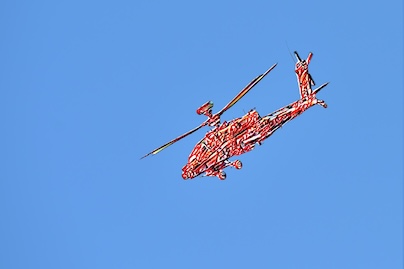}
        &
        \includegraphics[width=0.15\linewidth, height=0.7in]{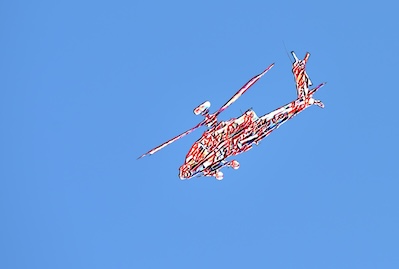}
        &
        \includegraphics[width=0.7in, height=0.7in]{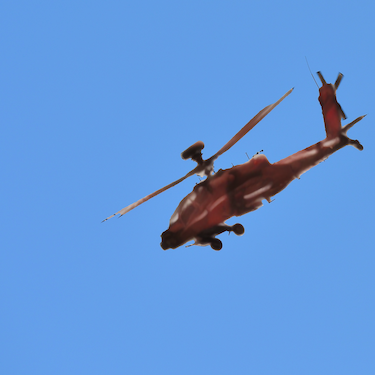}
        &
        \includegraphics[width=0.15\linewidth, height=0.7in]{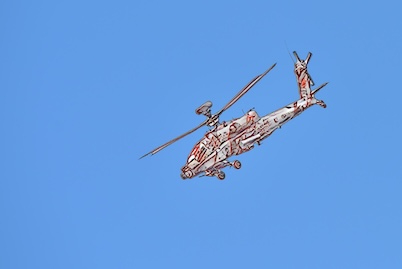}
        \\
        \includegraphics[width=0.15\linewidth, height=0.7in]{figures//results//More_Styled_Images/sa_224426.JPG}
        &
        \includegraphics[width=0.15\linewidth, height=0.7in]{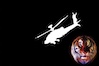}
        &
        \includegraphics[width=0.15\linewidth, height=0.7in]{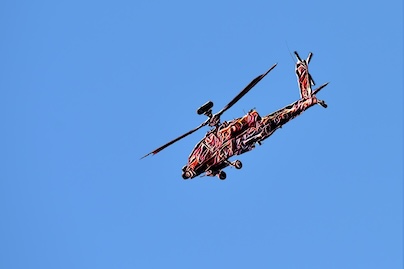}
        &
        \includegraphics[width=0.15\linewidth, height=0.7in]{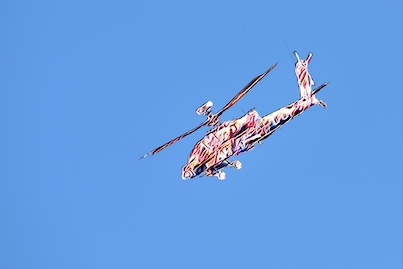}
        &
        \includegraphics[width=0.7in, height=0.7in]{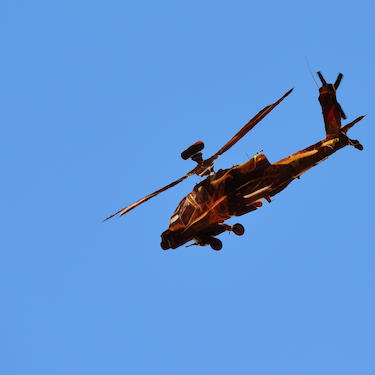}
        &
        \includegraphics[width=0.15\linewidth, height=0.7in]{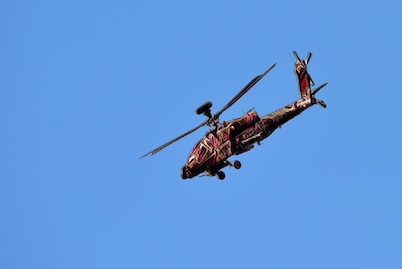}
        \\
        \includegraphics[width=0.15\linewidth, height=0.7in]{figures//results//More_Styled_Images/sa_224426.JPG}
        &
        \includegraphics[width=0.15\linewidth, height=0.7in]{figures//results//More_Styled_Images/helicopter8_mask_vs_style.JPG}
        &
        \includegraphics[width=0.15\linewidth, height=0.7in]{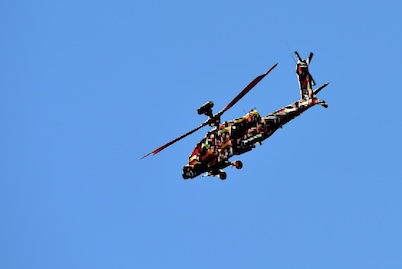}
        &
        \includegraphics[width=0.15\linewidth, height=0.7in]{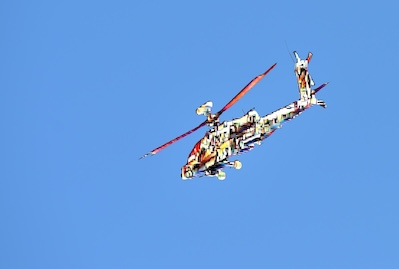}
        &
        \includegraphics[width=0.7in, height=0.7in]{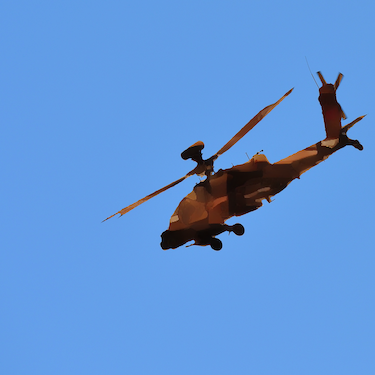}
        &
        \includegraphics[width=0.15\linewidth, height=0.7in]{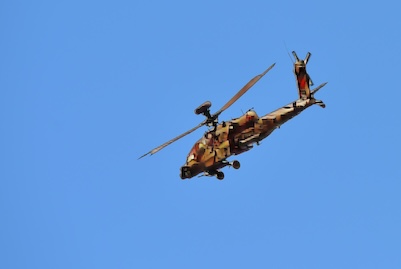}
        \\
        \includegraphics[width=0.15\linewidth, height=0.7in]{figures//results//More_Styled_Images/sa_224426.JPG}
        &
        \includegraphics[width=0.15\linewidth, height=0.7in]{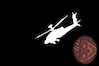}
        &
        \includegraphics[width=0.15\linewidth, height=0.7in]{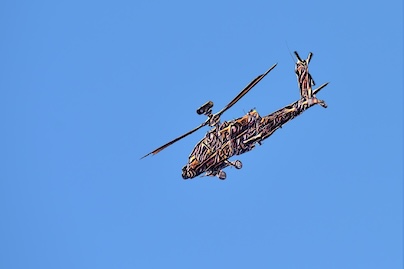}
        &
        \includegraphics[width=0.15\linewidth, height=0.7in]{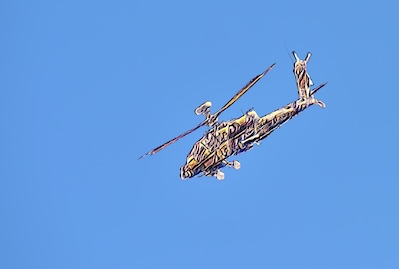}
        &
        \includegraphics[width=0.7in, height=0.7in]{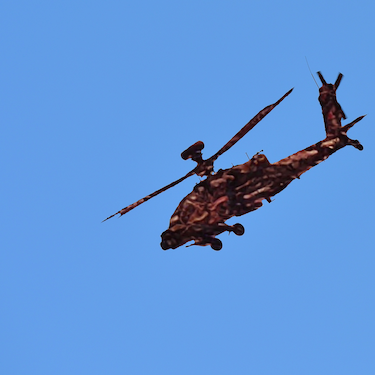}
        &
        \includegraphics[width=0.15\linewidth, height=0.7in]{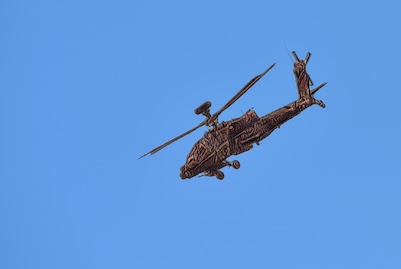}
        \\
        \includegraphics[width=0.15\linewidth, height=0.7in]{figures//results//More_Styled_Images/sa_224426.JPG}
        &
        \includegraphics[width=0.15\linewidth, height=0.7in]{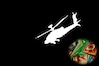}
        &
        \includegraphics[width=0.15\linewidth, height=0.7in]{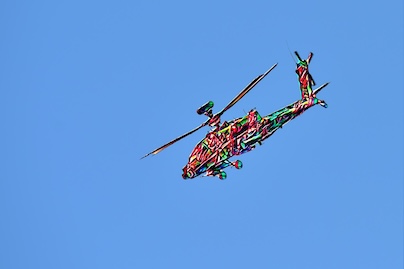}
        &
        \includegraphics[width=0.15\linewidth, height=0.7in]{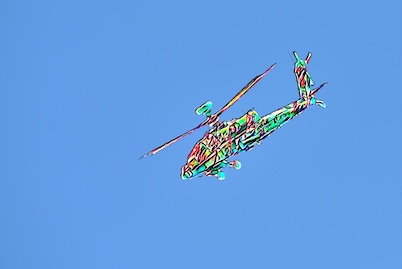}
        &
        \includegraphics[width=0.7in, height=0.7in]{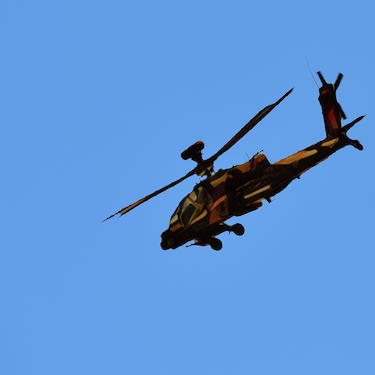}
        &
        \includegraphics[width=0.15\linewidth, height=0.7in]{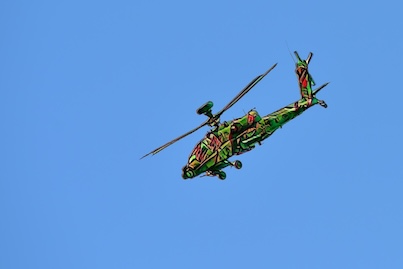}
        \\
        \includegraphics[width=0.15\linewidth, height=0.7in]{figures//results//More_Styled_Images/sa_224426.JPG}
        & 
        \includegraphics[width=0.15\linewidth, height=0.7in]{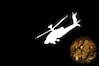}
        &
        \includegraphics[width=0.15\linewidth, height=0.7in]{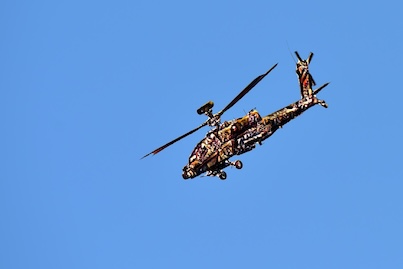}
        &
        \includegraphics[width=0.15\linewidth, height=0.7in]{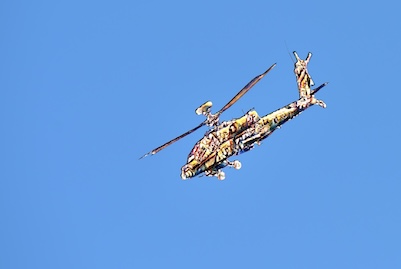}
        &
        \includegraphics[width=0.7in, height=0.7in]{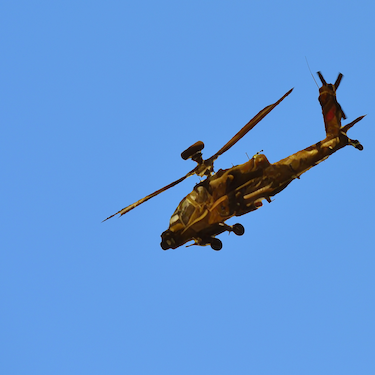}
        &
        \includegraphics[width=0.15\linewidth, height=0.7in]{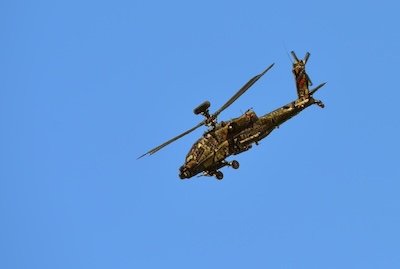}
        \\

        \includegraphics[width=0.15\linewidth, height=0.7in]{figures//results//More_Styled_Images/sa_224426.JPG}
        &
        \includegraphics[width=0.15\linewidth, height=0.7in]{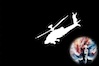}
        &
        \includegraphics[width=0.15\linewidth, height=0.7in]{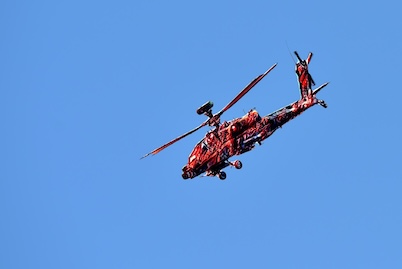}
        &
        \includegraphics[width=0.15\linewidth, height=0.7in]{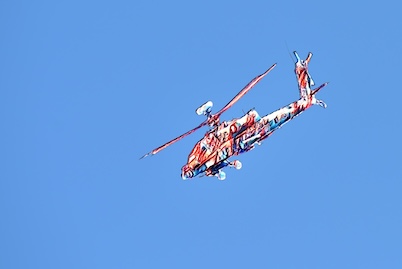}
        &
        \includegraphics[width=0.7in, height=0.7in]{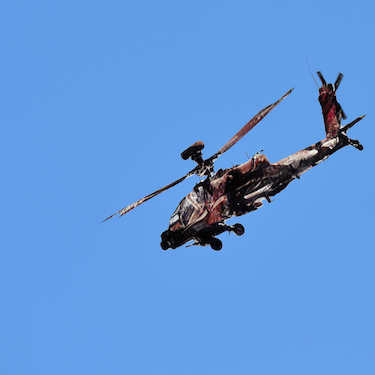}
        &
        \includegraphics[width=0.15\linewidth, height=0.7in]{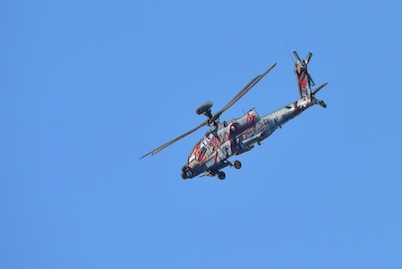}

        \\
        Original&
        Mask vs Style &
        Style-then-mask &
        Mask-then-style &
        StyleID &
        Ours \\
    \end{tabular}
\end{center}
   \caption{Helicopter example outputs from different style techniques.} 
\label{fig:results-grid-1}
\end{figure*}

\begin{figure*}[t]
\begin{center}
       \begin{tabular}{ccccccc}
        \includegraphics[width=0.15\linewidth, height=0.7in]{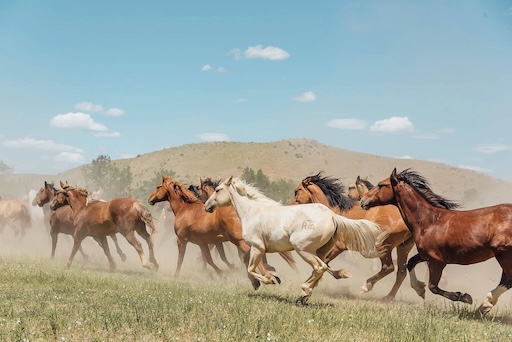}
        &
        \includegraphics[width=0.15\linewidth, height=0.7in]{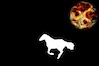}
        &
        \includegraphics[width=0.15\linewidth, height=0.7in]{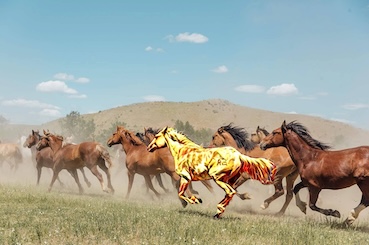}
        &
        \includegraphics[width=0.15\linewidth, height=0.7in]{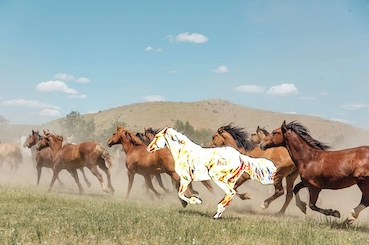}
        &
        \includegraphics[width=0.7in, height=0.7in]{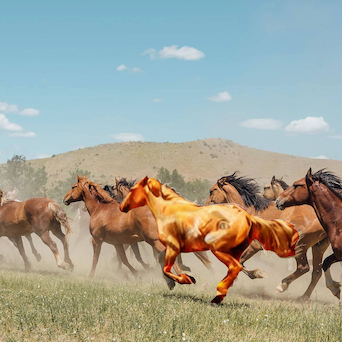}
        &
        \includegraphics[width=0.15\linewidth, height=0.7in]{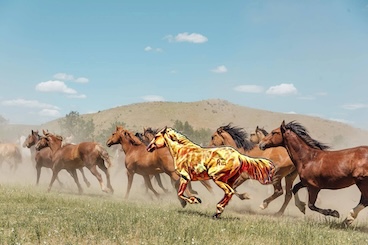}
        \\

        \includegraphics[width=0.15\linewidth, height=0.7in]{figures//results//More_Styled_Images/GettyImages_1207721867.JPG}
        &
        \includegraphics[width=0.15\linewidth, height=0.7in]{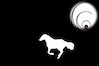}
        &
        \includegraphics[width=0.15\linewidth, height=0.7in]{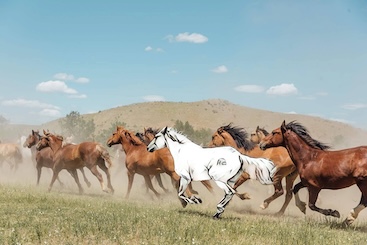}
        &
        \includegraphics[width=0.15\linewidth, height=0.7in]{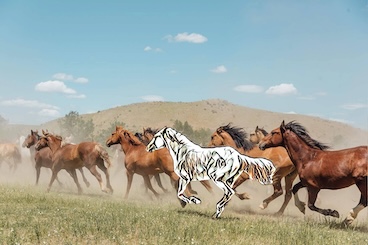}
        &
        \includegraphics[width=0.7in, height=0.7in]{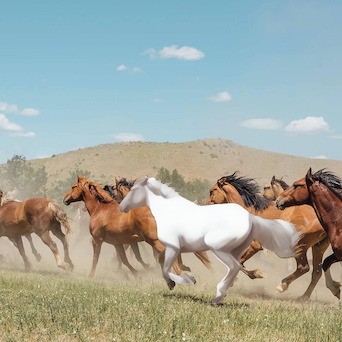}
        &
        \includegraphics[width=0.15\linewidth, height=0.7in]{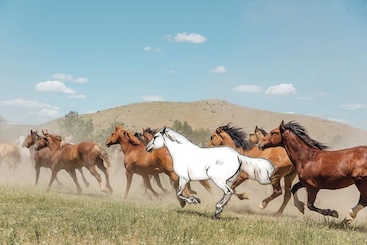}
        \\
        \includegraphics[width=0.15\linewidth, height=0.7in]{figures//results//More_Styled_Images/GettyImages_1207721867.JPG}
        &
        \includegraphics[width=0.15\linewidth, height=0.7in]{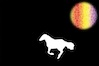}
        &
        \includegraphics[width=0.15\linewidth, height=0.7in]{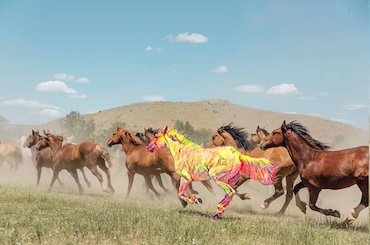}
        &
        \includegraphics[width=0.15\linewidth, height=0.7in]{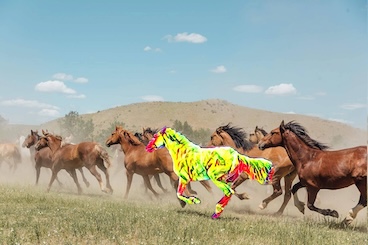}
        &
        \includegraphics[width=0.7in, height=0.7in]{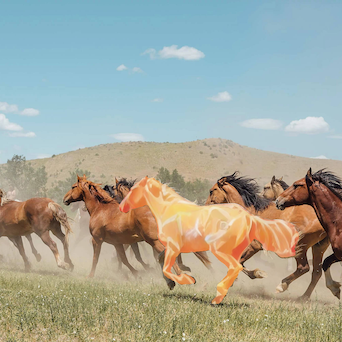}
        &
        \includegraphics[width=0.15\linewidth, height=0.7in]{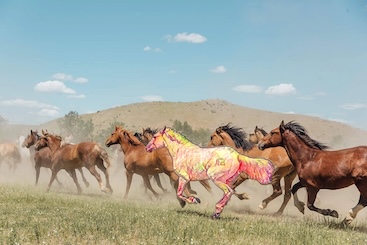}
        \\
        \includegraphics[width=0.15\linewidth, height=0.7in]{figures//results//More_Styled_Images/GettyImages_1207721867.JPG}
        &
        \includegraphics[width=0.15\linewidth, height=0.7in]{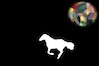}
        &
        \includegraphics[width=0.15\linewidth, height=0.7in]{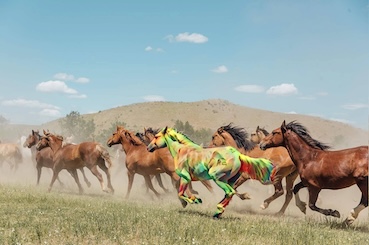}
        &
        \includegraphics[width=0.15\linewidth, height=0.7in]{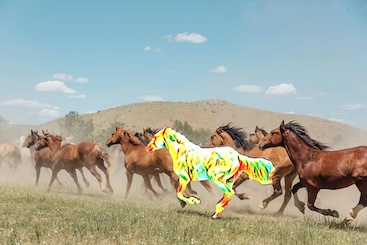}
        &
        \includegraphics[width=0.7in, height=0.7in]{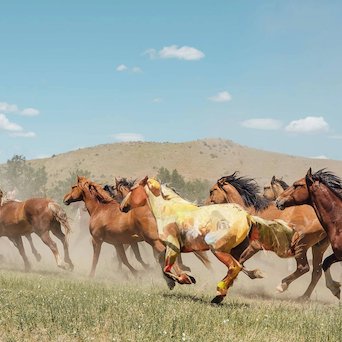}
        &
        \includegraphics[width=0.15\linewidth, height=0.7in]{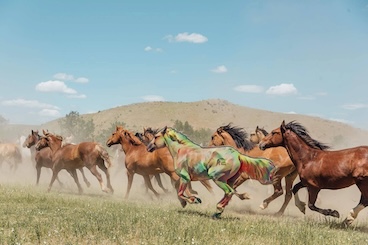}
        \\
        \includegraphics[width=0.15\linewidth, height=0.7in]{figures//results//More_Styled_Images/GettyImages_1207721867.JPG}
        &
        \includegraphics[width=0.15\linewidth, height=0.7in]{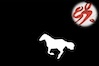}
        &
        \includegraphics[width=0.15\linewidth, height=0.7in]{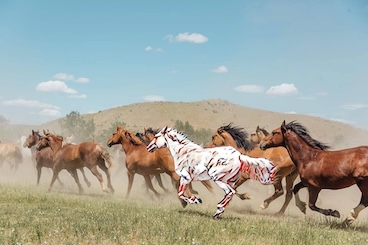}
        &
        \includegraphics[width=0.15\linewidth, height=0.7in]{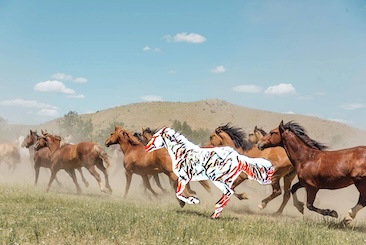}
        &
        \includegraphics[width=0.7in, height=0.7in]{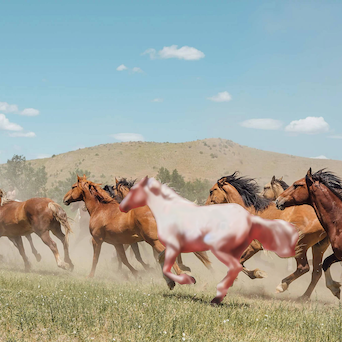}
        &
        \includegraphics[width=0.15\linewidth, height=0.7in]{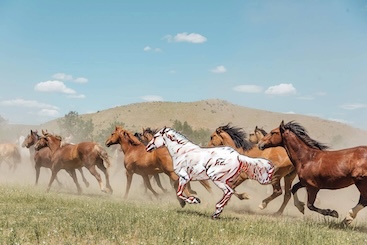}
        \\
        \includegraphics[width=0.15\linewidth, height=0.7in]{figures//results//More_Styled_Images/GettyImages_1207721867.JPG}
        &
        \includegraphics[width=0.15\linewidth, height=0.7in]{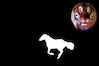}
        &
        \includegraphics[width=0.15\linewidth, height=0.7in]{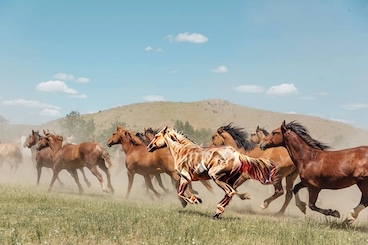}
        &
        \includegraphics[width=0.15\linewidth, height=0.7in]{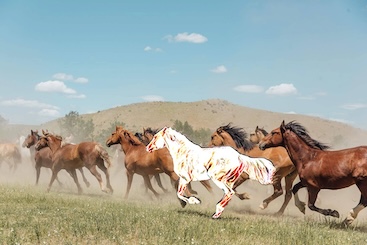}
        &
        \includegraphics[width=0.7in, height=0.7in]{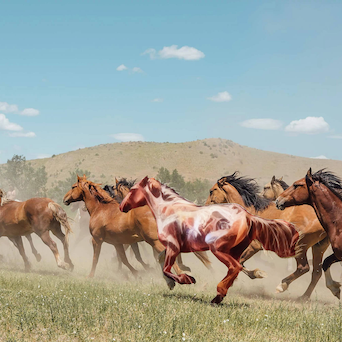}
        &
        \includegraphics[width=0.15\linewidth, height=0.7in]{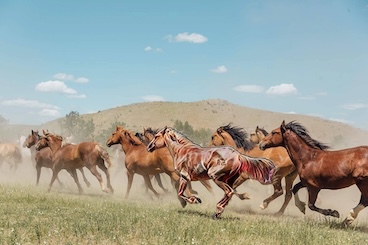}
        \\
        \includegraphics[width=0.15\linewidth, height=0.7in]{figures//results//More_Styled_Images/GettyImages_1207721867.JPG}
        &
        \includegraphics[width=0.15\linewidth, height=0.7in]{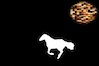}
        &
        \includegraphics[width=0.15\linewidth, height=0.7in]{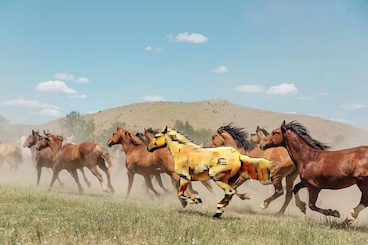}
        &
        \includegraphics[width=0.15\linewidth, height=0.7in]{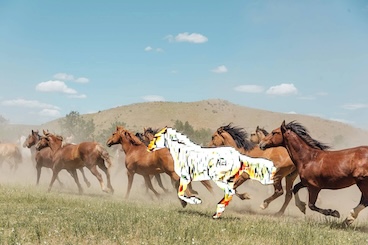}
        &
        \includegraphics[width=0.7in, height=0.7in]{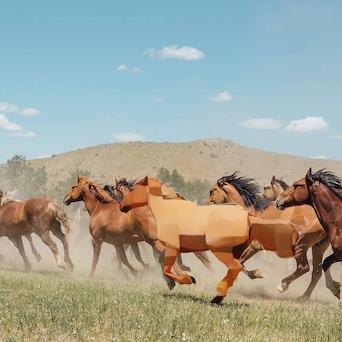}
        &
        \includegraphics[width=0.15\linewidth, height=0.7in]{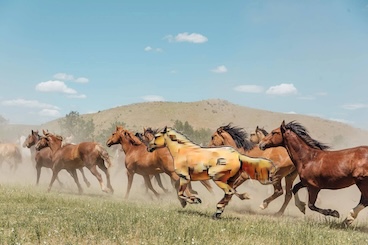}
        \\
        \includegraphics[width=0.15\linewidth, height=0.7in]{figures//results//More_Styled_Images/GettyImages_1207721867.JPG}
        &
        \includegraphics[width=0.15\linewidth, height=0.7in]{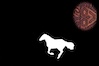}
        &
        \includegraphics[width=0.15\linewidth, height=0.7in]{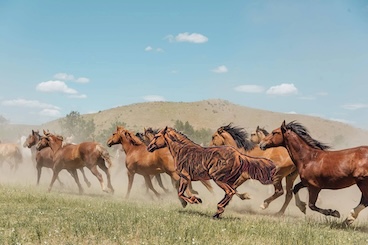}
        &
        \includegraphics[width=0.15\linewidth, height=0.7in]{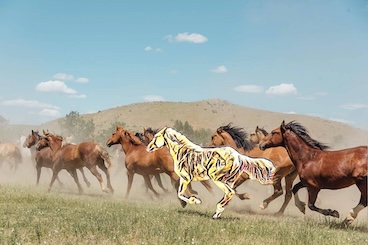}
        &
        \includegraphics[width=0.7in, height=0.7in]{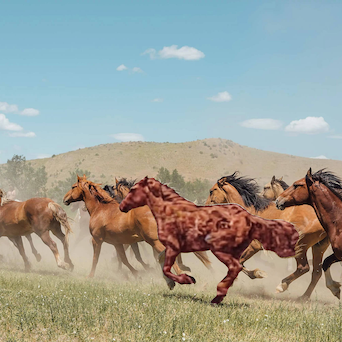}
        &
        \includegraphics[width=0.15\linewidth, height=0.7in]{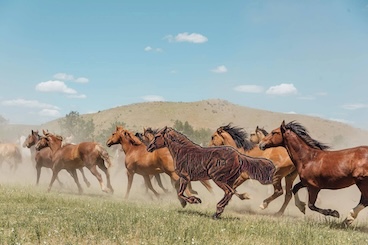}
        \\
        \includegraphics[width=0.15\linewidth, height=0.7in]{figures//results//More_Styled_Images/GettyImages_1207721867.JPG}
        &
        \includegraphics[width=0.15\linewidth, height=0.7in]{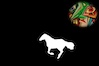}
        &
        \includegraphics[width=0.15\linewidth, height=0.7in]{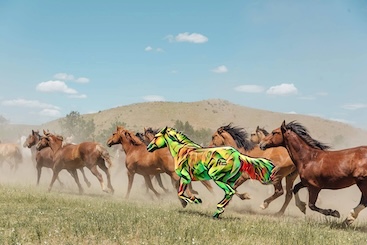}
        &
        \includegraphics[width=0.15\linewidth, height=0.7in]{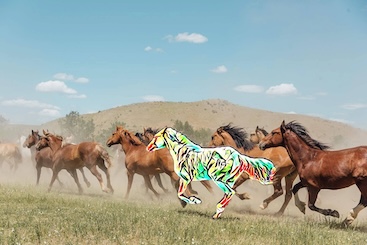}
        &
        \includegraphics[width=0.7in, height=0.7in]{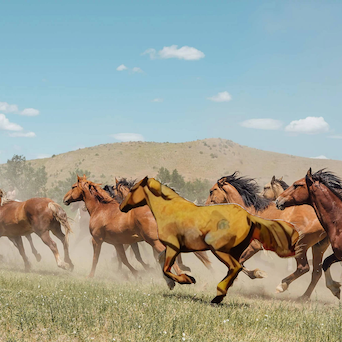}
        &
        \includegraphics[width=0.15\linewidth, height=0.7in]{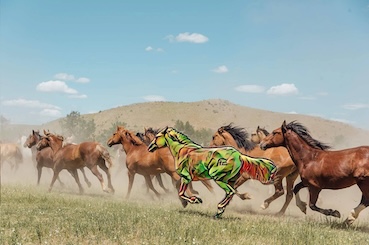}
        \\
        \includegraphics[width=0.15\linewidth, height=0.7in]{figures//results//More_Styled_Images/GettyImages_1207721867.JPG}
        &
        \includegraphics[width=0.15\linewidth, height=0.7in]{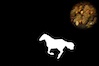}
        &
        \includegraphics[width=0.15\linewidth, height=0.7in]{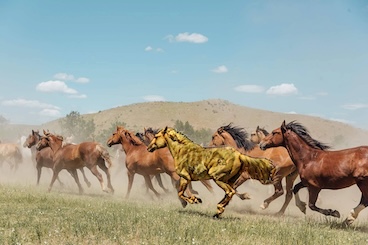}
        &
        \includegraphics[width=0.15\linewidth, height=0.7in]{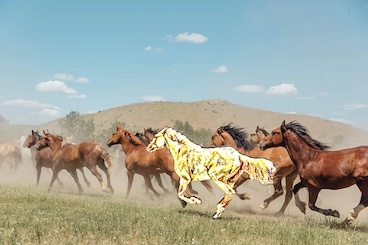}
        &
        \includegraphics[width=0.7in, height=0.7in]{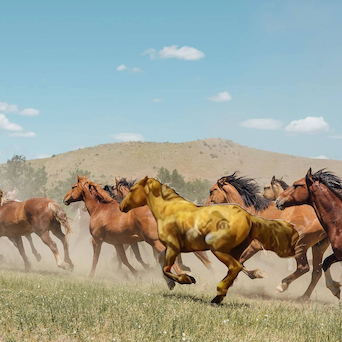}
        &
        \includegraphics[width=0.15\linewidth, height=0.7in]{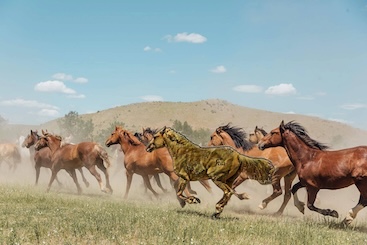}
        \\
        \includegraphics[width=0.15\linewidth, height=0.7in]{figures//results//More_Styled_Images/GettyImages_1207721867.JPG}
        &
        \includegraphics[width=0.15\linewidth, height=0.7in]{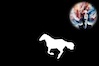}
        &
        \includegraphics[width=0.15\linewidth, height=0.7in]{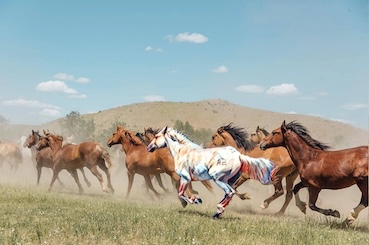}
        &
        \includegraphics[width=0.15\linewidth, height=0.7in]{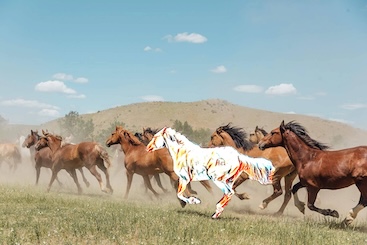}
        &
        \includegraphics[width=0.7in, height=0.7in]{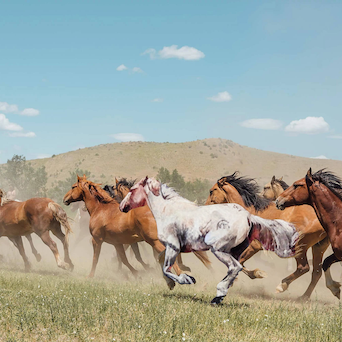}
        &
        \includegraphics[width=0.15\linewidth, height=0.7in]{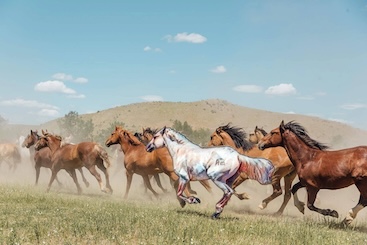}
        \\
        Original&
        Mask vs Style &
        Style-then-mask  &
        Mask-then-style &
        StyleID &
        Ours \\
    \end{tabular}
\end{center}
   \caption{Horse example outputs from different style techniques.} 
\label{fig:results-grid-2}
\end{figure*}

\begin{figure*}[t]
\begin{center}
       \begin{tabular}{cccccc}

        \includegraphics[width=0.15\linewidth, height=0.7in]{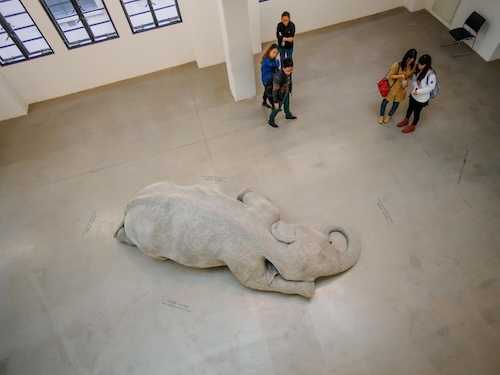}
        &
        \includegraphics[width=0.15\linewidth, height=0.7in]{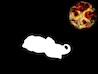}
        &
        \includegraphics[width=0.15\linewidth, height=0.7in]{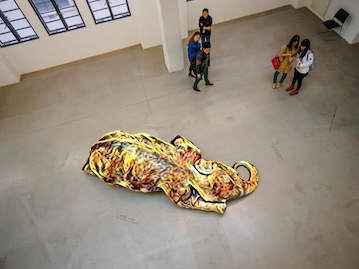}
        &
        \includegraphics[width=0.15\linewidth, height=0.7in]{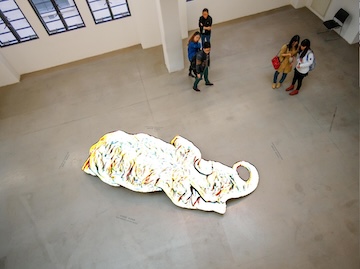}
        &
        \includegraphics[width=0.7in, height=0.7in]{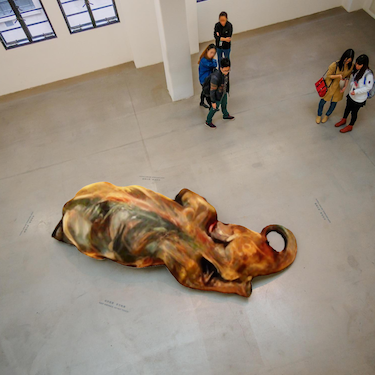}
        &
        \includegraphics[width=0.15\linewidth, height=0.7in]{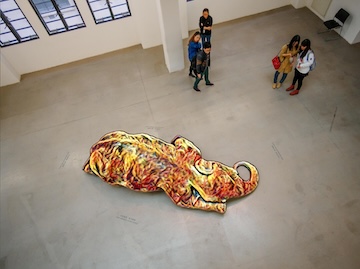}
        \\
        \includegraphics[width=0.15\linewidth, height=0.7in]{figures//results//More_Styled_Images/sa_224011.JPG}
        &
        \includegraphics[width=0.15\linewidth, height=0.7in]{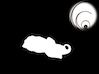}
        &
        \includegraphics[width=0.15\linewidth, height=0.7in]{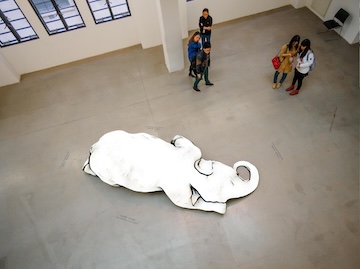}
        &
        \includegraphics[width=0.15\linewidth, height=0.7in]{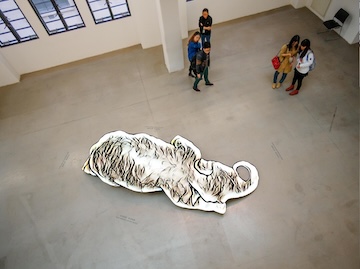}
        &
        \includegraphics[width=0.7in, height=0.7in]{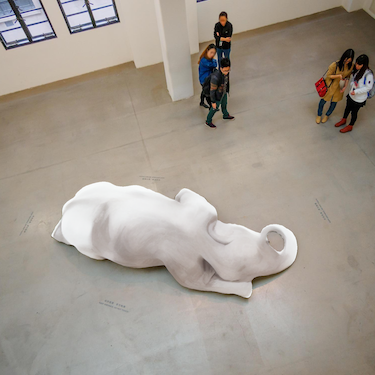}
        &
        \includegraphics[width=0.15\linewidth, height=0.7in]{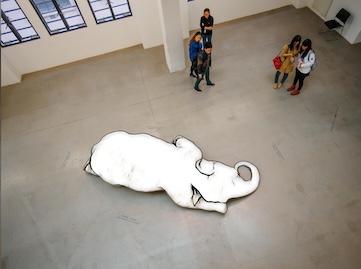}
        \\
        \includegraphics[width=0.15\linewidth, height=0.7in]{figures//results//More_Styled_Images/sa_224011.JPG}
        &
        \includegraphics[width=0.15\linewidth, height=0.7in]{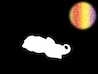}
        &
        \includegraphics[width=0.15\linewidth, height=0.7in]{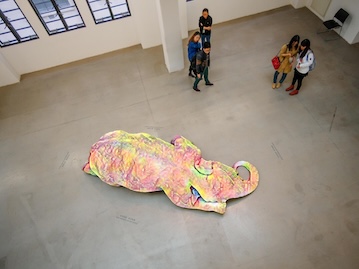}
        &
        \includegraphics[width=0.15\linewidth, height=0.7in]{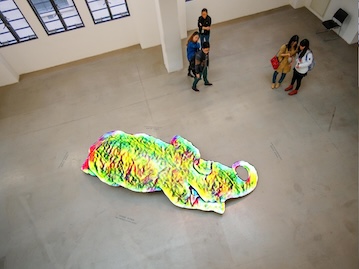}
        &
        \includegraphics[width=0.7in, height=0.7in]{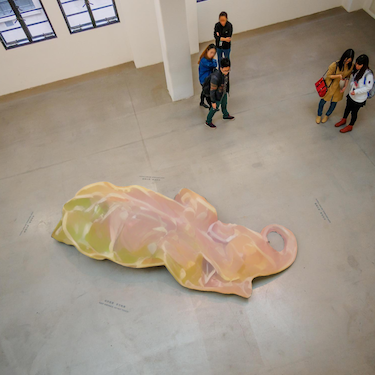}
        &
        \includegraphics[width=0.15\linewidth, height=0.7in]{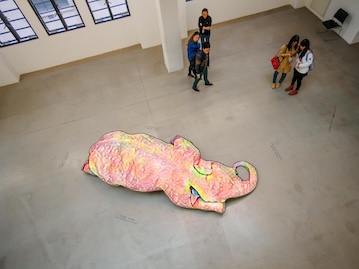}
        \\
        \includegraphics[width=0.15\linewidth, height=0.7in]{figures//results//More_Styled_Images/sa_224011.JPG}
        &
        \includegraphics[width=0.15\linewidth, height=0.7in]{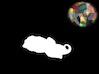}
        &
        \includegraphics[width=0.15\linewidth, height=0.7in]{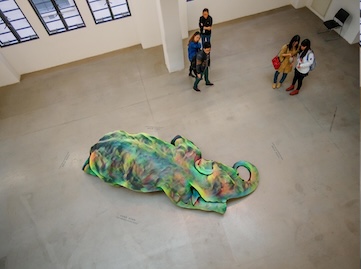}
        &
        \includegraphics[width=0.15\linewidth, height=0.7in]{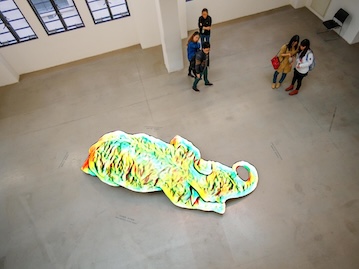}
        &
        \includegraphics[width=0.7in, height=0.7in]{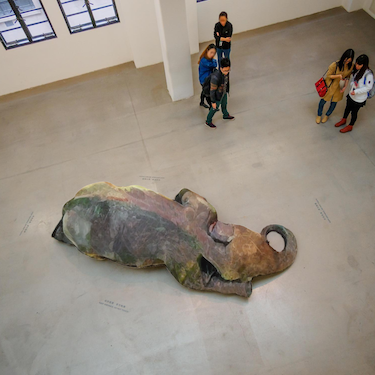}
        &
        \includegraphics[width=0.15\linewidth, height=0.7in]{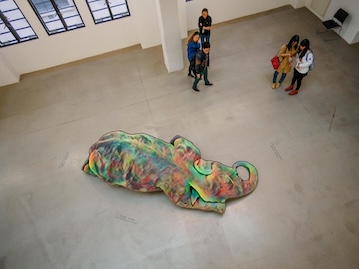}
        \\
        \includegraphics[width=0.15\linewidth, height=0.7in]{figures//results//More_Styled_Images/sa_224011.JPG}
        &
        \includegraphics[width=0.15\linewidth, height=0.7in]{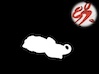}
        &
        \includegraphics[width=0.15\linewidth, height=0.7in]{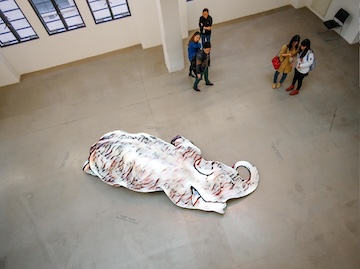}
        &
        \includegraphics[width=0.15\linewidth, height=0.7in]{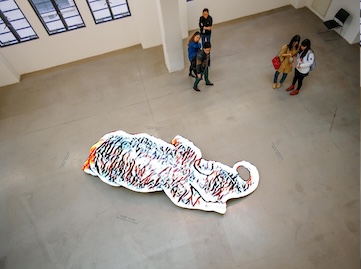}
        &
        \includegraphics[width=0.7in, height=0.7in]{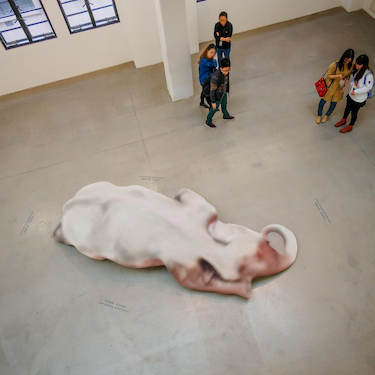}
        &
        \includegraphics[width=0.15\linewidth, height=0.7in]{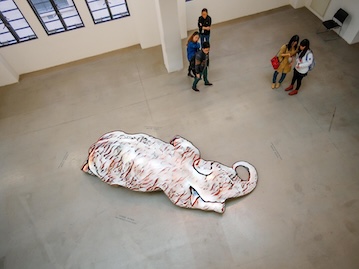}
        \\
        \includegraphics[width=0.15\linewidth, height=0.7in]{figures//results//More_Styled_Images/sa_224011.JPG}
        &
        \includegraphics[width=0.15\linewidth, height=0.7in]{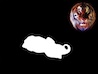}
        &
        \includegraphics[width=0.15\linewidth, height=0.7in]{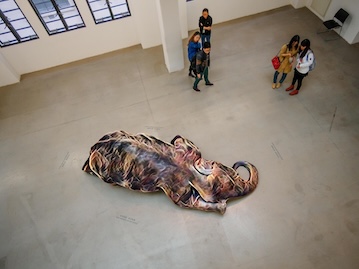}
        &
        \includegraphics[width=0.15\linewidth, height=0.7in]{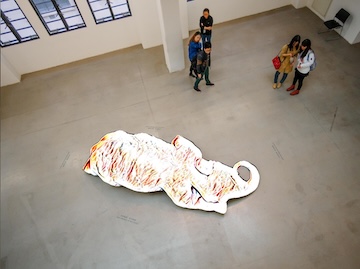}
        &
        \includegraphics[width=0.7in, height=0.7in]{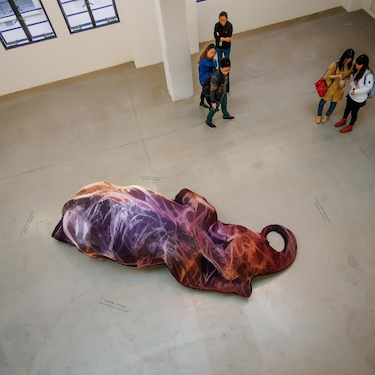}
        &
        \includegraphics[width=0.15\linewidth, height=0.7in]{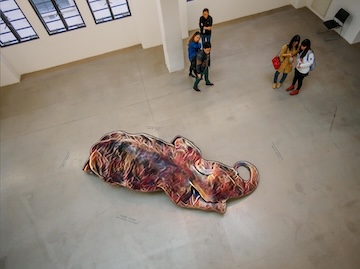}
        \\
        \includegraphics[width=0.15\linewidth, height=0.7in]{figures//results//More_Styled_Images/sa_224011.JPG}
        &
        \includegraphics[width=0.15\linewidth, height=0.7in]{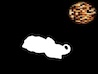}
        &
        \includegraphics[width=0.15\linewidth, height=0.7in]{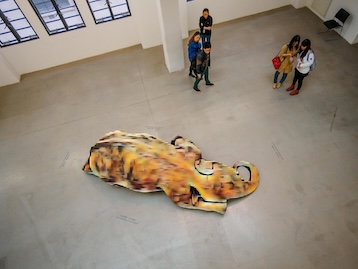}
        &
        \includegraphics[width=0.15\linewidth, height=0.7in]{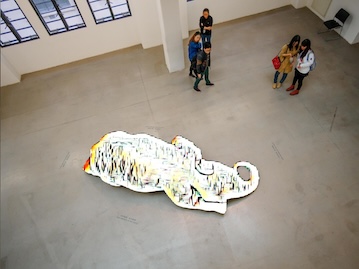}
        &
        \includegraphics[width=0.7in, height=0.7in]{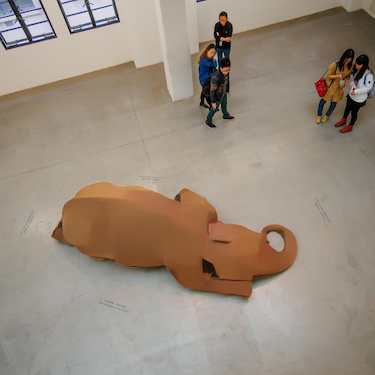}
        &
        \includegraphics[width=0.15\linewidth, height=0.7in]{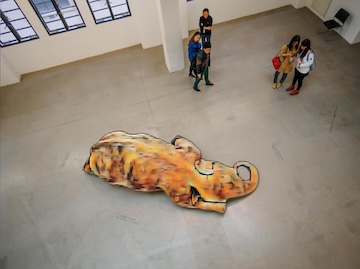}
        \\
        \includegraphics[width=0.15\linewidth, height=0.7in]{figures//results//More_Styled_Images/sa_224011.JPG}
        &
        \includegraphics[width=0.15\linewidth, height=0.7in]{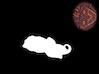}
        &
        \includegraphics[width=0.15\linewidth, height=0.7in]{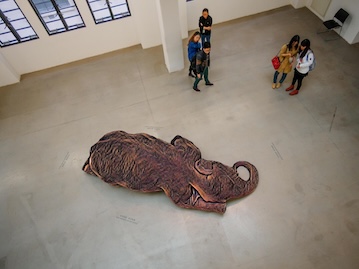}
        &
        \includegraphics[width=0.15\linewidth, height=0.7in]{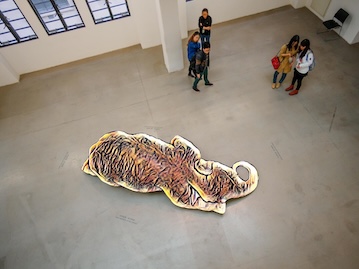}
        &
        \includegraphics[width=0.7in, height=0.7in]{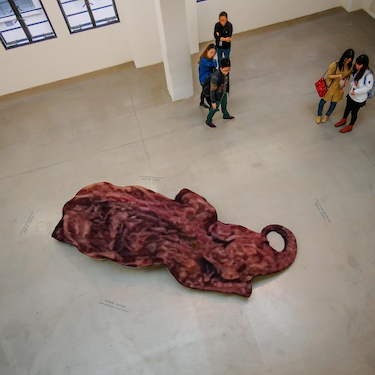}
        &
        \includegraphics[width=0.15\linewidth, height=0.7in]{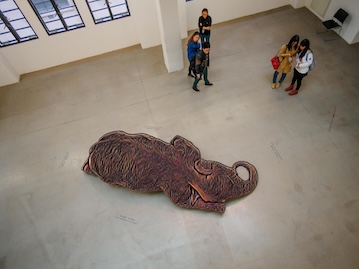}
        \\
        \includegraphics[width=0.15\linewidth, height=0.7in]{figures//results//More_Styled_Images/sa_224011.JPG}
        &
        \includegraphics[width=0.15\linewidth, height=0.7in]{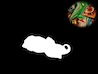}
        &
        \includegraphics[width=0.15\linewidth, height=0.7in]{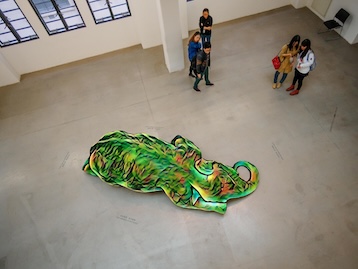}
        &
        \includegraphics[width=0.15\linewidth, height=0.7in]{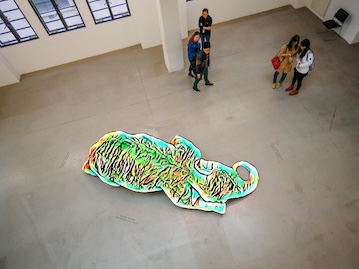}
        &
        \includegraphics[width=0.7in, height=0.7in]{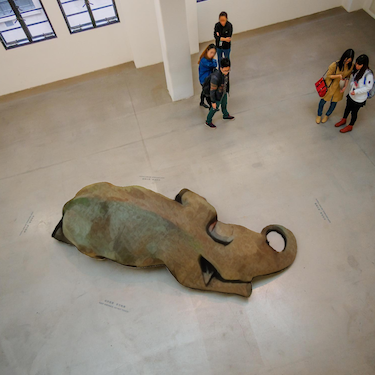}
        &
        \includegraphics[width=0.15\linewidth, height=0.7in]{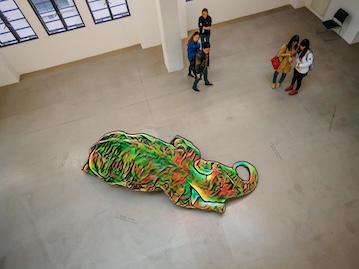}
        \\
        \includegraphics[width=0.15\linewidth, height=0.7in]{figures//results//More_Styled_Images/sa_224011.JPG}
        &
        \includegraphics[width=0.15\linewidth, height=0.7in]{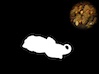}
        &
        \includegraphics[width=0.15\linewidth, height=0.7in]{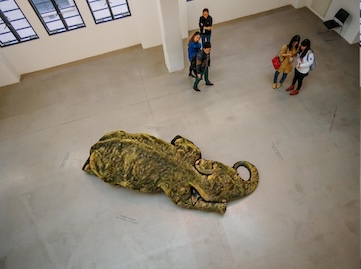}
        &
        \includegraphics[width=0.15\linewidth, height=0.7in]{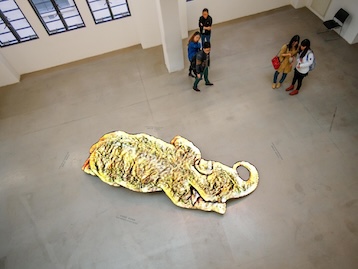}
        &
        \includegraphics[width=0.7in, height=0.7in]{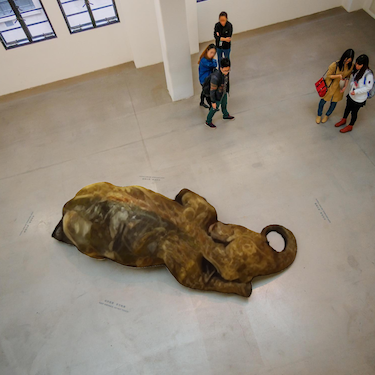}
        &
        \includegraphics[width=0.15\linewidth, height=0.7in]{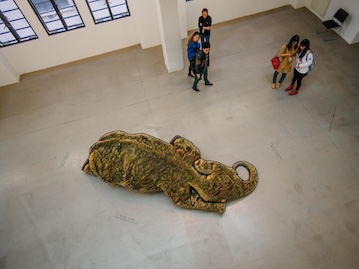}
        \\
        \includegraphics[width=0.15\linewidth, height=0.7in]{figures//results//More_Styled_Images/sa_224011.JPG}
        &
        \includegraphics[width=0.15\linewidth, height=0.7in]{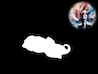}
        &
        \includegraphics[width=0.15\linewidth, height=0.7in]{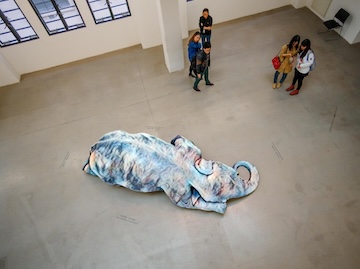}
        &
        \includegraphics[width=0.15\linewidth, height=0.7in]{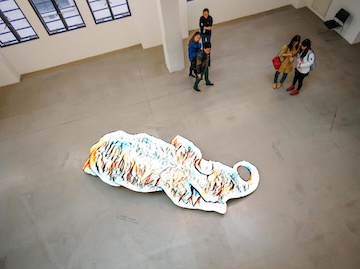}
        &
        \includegraphics[width=0.7in, height=0.7in]{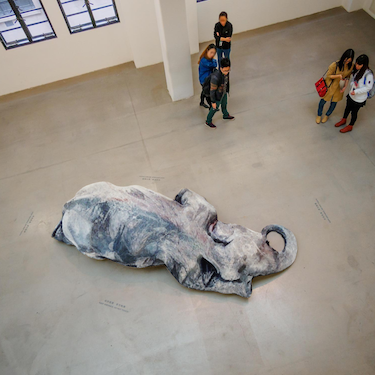}
        &
        \includegraphics[width=0.15\linewidth, height=0.7in]{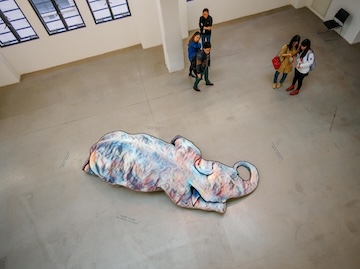}
        \\

        Original&
        Mask vs Style &
        Style-then-mask &
        Mask-then-style &
        StyleID &
        Ours \\
    \end{tabular}
\end{center}
   \caption{Elephant example outputs from different style techniques.} 
\label{fig:results-grid-3}
\end{figure*}

Figs.~\ref{fig:results-grid-1}, \ref{fig:results-grid-2}, and \ref{fig:results-grid-3} provide additional comparisons between our method and other techniques.

